\documentclass[acmlarge, nonacm]{acmart}

\usepackage{lipsum}
\usepackage{graphicx}
\usepackage{lscape}
\usepackage{tabularx}
\usepackage{booktabs}
\usepackage{xltabular}
\usepackage{caption}
\usepackage{subcaption}
\usepackage{tikz}
\usepackage{listings}
\usepackage{fontawesome}
\usepackage{tablefootnote}
\usepackage[edges]{forest}
\usepackage[super]{nth}

\usepackage[T1]{fontenc}
\usepackage[utf8]{inputenc}
\usepackage[english]{babel}
\usepackage[autostyle=true]{csquotes}
\usepackage[nolist,nohyperlinks]{acronym}

\newcolumntype{Y}{>{\raggedleft\arraybackslash}X}
\newcolumntype{R}{>{\hsize=0.6\hsize}Y}
\newcolumntype{S}{>{\hsize=0.2\hsize}Y}
\newcolumntype{T}{>{\hsize=0.3\hsize}Y}
\newcolumntype{U}{>{\hsize=0.175\hsize}Y}

\newcommand{\deart}{DEArt}
\newcommand{\iart}{iART}
\newcommand{\peopleart}{People-Art}
\newcommand{\popart}{PoPArt}
\newcommand{\wikiart}{WikiArt}

\defcitealias{madhu2020b}{2020}

\makeatletter
\@addtoreset{subfigure}{row}
\makeatother

\makeatletter
\renewcommand\p@subfigure{Fig.~\thefigure}
\renewcommand\p@figure{Fig.~}
\renewcommand\p@table{Table~}
\makeatother

\let\origthelstnumber\thelstnumber
\makeatletter

\newcommand*\stopnumber
{
    \lst@AddToHook{OnNewLine}
    {
        \let\thelstnumber\relax
        \advance\c@lstnumber-\@ne\relax
    }
}

\newcommand*\startnumber[1]
{
    \setcounter{lstnumber}{\numexpr#1-1\relax}
    \lst@AddToHook{OnNewLine}
    {
        \let\thelstnumber\origthelstnumber
        \refstepcounter{lstnumber}
    }
}
\makeatother

\definecolor{BurntOrange}{HTML}{F7921D}

\lstdefinelanguage{json}{
    escapeinside={(*@}{@*)},
    basicstyle=\fontseries{l}\fontfamily{lmr}\scriptsize\ttfamily,
    commentstyle=\color[HTML]{006EB8}, 
    stringstyle=\color[HTML]{00A64F}, 
    keywordstyle=\itshape\color[HTML]{ED1B23},
    numberstyle=\tiny,
    stepnumber=1,
    numbersep=8pt,
    showstringspaces=false,
    breaklines=true,
    string=[s]{"}{"},
    comment=[l]{:\ "},
    morecomment=[l]{:"},
    keywords=[2]{true,false},
    literate=
        *{0}{{{\color{BurntOrange}0}}}{1}
         {1}{{{\color{BurntOrange}1}}}{1}
         {2}{{{\color{BurntOrange}2}}}{1}
         {3}{{{\color{BurntOrange}3}}}{1}
         {4}{{{\color{BurntOrange}4}}}{1}
         {5}{{{\color{BurntOrange}5}}}{1}
         {6}{{{\color{BurntOrange}6}}}{1}
         {7}{{{\color{BurntOrange}7}}}{1}
         {8}{{{\color{BurntOrange}8}}}{1}
         {9}{{{\color{BurntOrange}9}}}{1}
}

\AtBeginDocument{%
  \providecommand\BibTeX{{%
    \normalfont B\kern-0.5em{\scshape i\kern-0.25em b}\kern-0.8em\TeX}}}

\setcopyright{none}

\begin{document}


\begin{acronym}
\acro{AP}{Average Precision}
\acro{AR}{Average Recall}
\acro{CLIP}{Contrastive Language-image Pre-training}
\acro{COCO}{Common Objects in Context}
\acro{CNN}{Convolutional Neural Network}
\acro{EMA}{Exponential Moving Average}
\acro{GLAM}{Galleries, Libraries, Archives, and Museums}
\acro{IoU}{Intersection over Union}
\acro{PVT}{Pyramid Vision Transformer}
\acro{R-CNN}{Region-based Convolutional Neural Network}
\acro{RPN}{Region Proposal Network}
\acro{SGD}{Stochastic Gradient Descent}
\acro{TOOD}{Task-aligned One-stage Object Detection}
\acro{UMAP}{Uniform Manifold Approximation and Projection}
\acro{VOC}{Visual Object Classes}
\end{acronym}

\title[Poses of People in Art]{Poses of People in Art: A Data Set for Human Pose Estimation in Digital Art History}

\author{Stefanie Schneider}
\email{stefanie.schneider@itg.uni-muenchen.de}
\orcid{0000-0003-4915-6949}
\author{Ricarda Vollmer}
\email{ricarda.vollmer@campus.lmu.de}
\orcid{0000-0002-4105-1045}
\affiliation{%
  \institution{Ludwig Maximilian University of Munich}
  \city{Geschwister-Scholl-Platz 1, Munich}
  \country{Germany}
}

\renewcommand{\shortauthors}{S. Schneider and R. Vollmer}

\begin{abstract}
Throughout the history of art, the pose---as the holistic abstraction of the human body's expression---has proven to be a constant in numerous studies. However, due to the enormous amount of data that so far had to be processed by hand, its crucial role to the formulaic recapitulation of art-historical motifs since antiquity could only be highlighted selectively. This is true even for the now automated estimation of human poses, as domain-specific, sufficiently large data sets required for training computational models are either not publicly available or not indexed at a fine enough granularity. With the \textit{Poses of People in Art} data set, we introduce the first openly licensed data set for estimating human poses in art and validating human pose estimators. It consists of $2{,}454$ images from $22$ art-historical depiction styles, including those that have increasingly turned away from lifelike representations of the body since the \nth{19} century. A total of $10{,}749$ human figures are precisely enclosed by rectangular bounding boxes, with a maximum of four per image labeled by up to $17$ keypoints; among these are mainly joints such as elbows and knees. For machine learning purposes, the data set is divided into three subsets---training, validation, and testing---, that follow the established JSON-based Microsoft \ac{COCO} format, respectively. Each image annotation, in addition to mandatory fields, provides metadata from the art-historical online encyclopedia \wikiart. With this paper, we elaborate on the acquisition and constitution of the data set, address various application scenarios, and discuss prospects for a digitally supported art history. We show that the data set enables the comprehensive investigation of body phenomena in art, whether at the level of individual figures, which can thus be captured in their subtleties, or entire figure constellations, whose position, distance, or proximity to one another is considered.
\end{abstract}



\settopmatter{printacmref=false, printccs=true, printfolios=true}

\maketitle
\acresetall

\section{Introduction}
\label{chp:introduction}

The abstracted human body, into which measurements, proportions, and movements are inscribed, has played a crucial role throughout the history of art. This particularly applies to the drawing apprenticeship \cite{rathgeber2011}, 
whose best-known example is Leonardo da Vinci's \textit{Vitruvian Man}. 
As early as the \nth{17} century, artists began to structure the human pose\footnote{For reasons of simplicity, we hereinafter do not distinguish between the terms \enquote*{posture} and \enquote*{pose.} Instead, we use the term \enquote*{pose} for any kind of bodily expression.} into a \enquote*{language} of non-verbal communication~\cite{knowlson1965}, pursued with scientific meticulousness into the \nth{18} century, e.g., by the Physiognomist Johann Caspar Lavater~\cite{fliedl1992}. 
Attempts to establish a kind of pose vocabulary, however, have been made primarily in relation to hand gestures~\cite{bulwer1644, amira1905}, with references to antiquity evident in most efforts~\cite{brilliant1963}. It was the Finnish art historian Johan Jakob Tikkanen who, in the \nth{19} century, then sought to motivate a differentiated terminology of leg positions~\cite{tikkanen1912}, drawing on perspectives from the natural sciences, such as Darwin's essays on the expression of humans and animals~\cite{darwin1877} as well as botanical classification systems~\cite{kauffmann1993}. In contrast, the studies of the art historian and cultural theorist Aby Warburg at the beginning of the \nth{20} century should not be understood as standardized~\cite{pfisterer2020}: 
through his concept of \enquote*{Pathosformeln,} Warburg rather loosely examined body phenomena recurring since antiquity~\cite{warburg1998a, warburg1998b}. 
 
This high selectivity of art-historical research---especially when compared to other body-oriented disciplines such as theater and dance studies~\cite{schlemmer1969, rathgeber2019}---can be attributed to various reasons. We perceive two factors as pivotal: (i)~the enormous amount of data that for a comprehensive analysis so far had to be processed by hand, and (ii)~the lack of an approach that holistically and systematically assesses human pose through relevant \textit{keypoints}, e.g., wrists or knees. With the ongoing digitization and online publication of historical objects, researchers could now potentially draw on increasingly large collections of images to examine dominant pose types or time-dependent body phenomena. To date, however, few approaches to automatically estimate human poses in art-historical imagery have emerged~\cite{impett2016, impett2017, jenicek2019, madhu2020a, madhu2020b}, possibly due to the lack of domain-specific, sufficiently large data sets required for training computational models, e.g., \acp{CNN}. Existing data sets fall broadly into two categories. Either they do index keypoints but are not publicly available and are dedicated to a comparatively narrow subset of art-historical representation practices~\cite{impett2016, madhu2020b}. Or they are freely accessible to the public but enclose human figures only by rectangular \textit{bounding boxes}; their pose is then broadly categorized without specifically delineating keypoints~\cite{deart}.

\begin{figure}
\centering
\includegraphics[width=\linewidth]{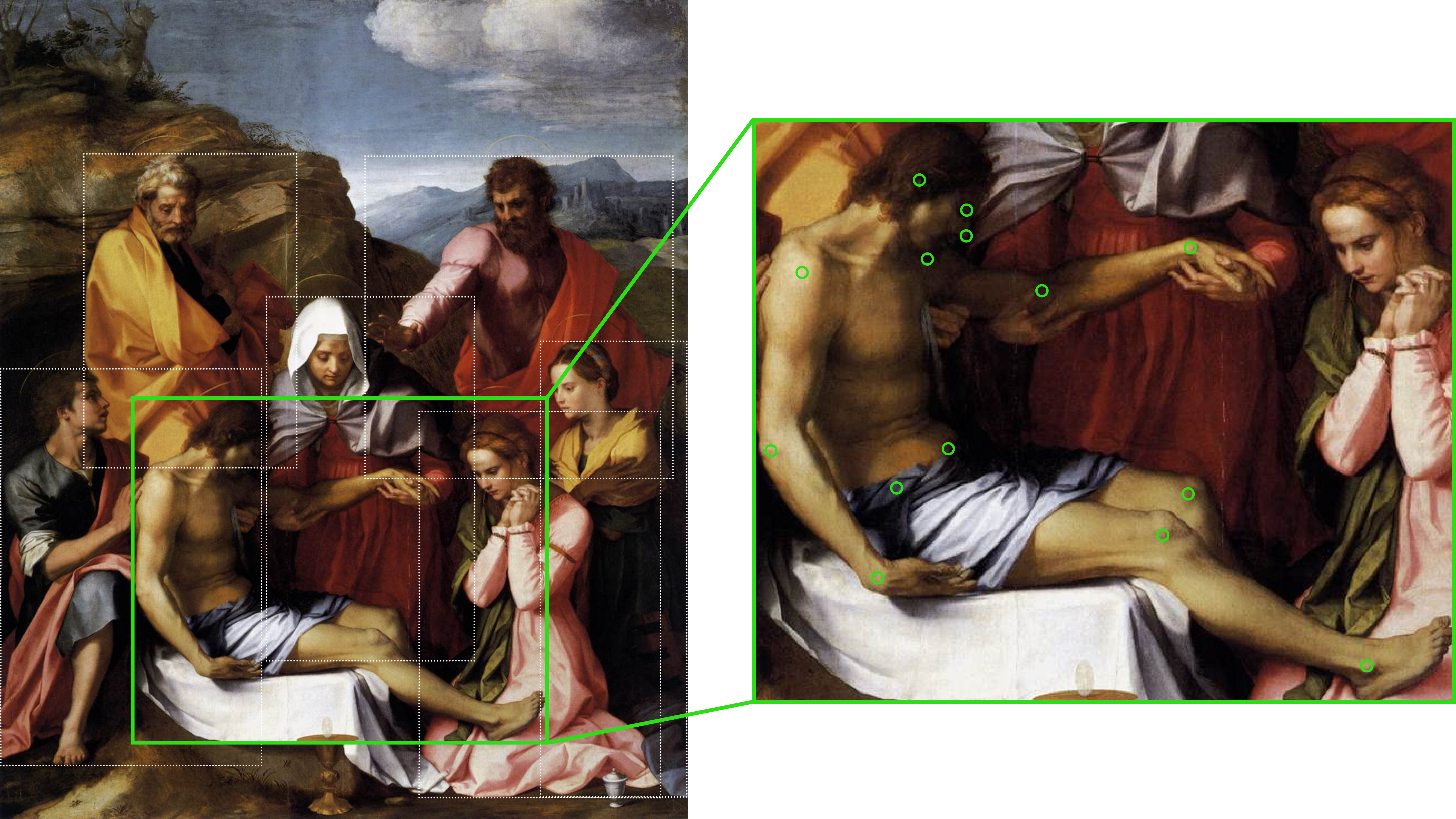}
\caption{
We differentiate between two annotation modes: bounding box and keypoint annotation. First, as shown on the left in Andrea del Sarto's \textit{Pietà with Saints} (1523--1524), human figures are marked with bounding boxes enclosing them. For a maximum of four per image, up to $17$ pose-relevant keypoints are then assigned, which are indicated with green circles in the detail view on the right.
}
\label{fig:overview-popart}
\end{figure}

Our contributions are three-fold. (i)~With \textit{Poses of People in Art}, hereinafter abbreviated to \popart, we introduce the first publicly available and openly licensed data set for estimating human poses in art. It is composed of $10{,}749$ bounding box and $56{,}154$ keypoint annotations from $22$ art-historical depiction styles, including those that have emerged since the \nth{19} century and have increasingly turned away from lifelike representations of the body; \ref{fig:overview-popart} illustrates both annotation modes. (ii)~We demonstrate that \popart\ enables the quantitatively systematized exploration of human pose in visual art by capturing the body holistically and across different stylistic periods. Pose may thus emerge as wholly elemental to the formulaic recapitulation of significant topoi and motifs through computational assistance. (iii)~As a by-product of \popart's domain-specific curation, the sole detection of figures in art-historical collections is decisively improved. In contrast to the similarly constituted \peopleart\ data set~\cite{westlake2016}, which also exclusively labels human figures, \popart\ contains fewer training, validation, and testing images. It, however, features nearly three times as many positive training samples with at least one figure instance annotation.

The remainder of this paper is structured as follows. In Section~\ref{chp:related-work}, we first review art-historically relevant data sets that can be leveraged for image classification and object detection tasks. Section~\ref{chp:data-set} then elaborates on the acquisition and constitution of the \popart\ data set. In this context, we also clarify the annotation guidelines we adapted to the domain. In the course of Section~\ref{chp:applications}, we address various application scenarios and discuss prospects for a digitally supported art history. Lastly, Section~\ref{chp:conclusion} concludes the paper and outlines areas for potential future research. The data set is available as a version-controlled repository on Zenodo.\footnote{\url{https://doi.org/10.5281/zenodo.7516230}.}

\begin{table}[t!]
\footnotesize
\begin{xltabular}{\columnwidth}{@{}Xl*{7}{U}@{}}
\caption{
Art-historically relevant data sets for image classification and object detection tasks are compared. Grey check marks specify information that is not directly stored in the respective data set, but has to be accessed via the referenced content providers.
}
\label{tab:related-data-sets}\\
\toprule
Name & Author(s) & Year & \multicolumn{2}{c}{Annotation} & \multicolumn{2}{c}{Levels} & \multicolumn{2}{c}{Availability}\\
 & & & \textit{Formal} & \textit{Content} & \textit{Image} & \textit{Object} & \textit{Public} & \textit{Privat}\\
\midrule
Medieval Manuscripts \cite{yarlagadda2010} & \citeauthor{yarlagadda2010} & \citeyear{yarlagadda2010} & \checkmark & \checkmark & \checkmark & $\checkmark^{1}$ & & \checkmark \\
\wikiart\ (f.k.a. WikiPaintings) \cite{wikiart} & Unknown & \citeyear{wikiart} & \checkmark & & \checkmark & & \checkmark & \\ 
PrintART \cite{printart} & \citeauthor{printart} & \citeyear{printart} & & \checkmark & \checkmark & $\checkmark^{1}$ & & \checkmark\\ 
Paintings \cite{paintings} & \citeauthor{paintings} & \citeyear{paintings} & $\color{lightgray}\checkmark$ & \checkmark & \checkmark & & \checkmark & \\ 
Picasso \cite{picasso} & \citeauthor{picasso} & \citeyear{picasso} & \checkmark & \checkmark & \checkmark & & & \checkmark\\
Painting-91 \cite{painting-91} & \citeauthor{painting-91} & \citeyear{painting-91} & \checkmark & & \checkmark & & & \checkmark\\ 
Rijksmuseum Challenge \cite{rijksmuseum-challenge} & \citeauthor{rijksmuseum-challenge} & \citeyear{rijksmuseum-challenge} & \checkmark & & \checkmark & & \checkmark & \\ 
Pandora \cite{pandora} & \citeauthor{pandora} & \citeyear{pandora} & \checkmark & & \checkmark & & & \checkmark \\
Warburg's Bilderatlas \cite{impett2016} & \citeauthor{impett2016} & \citeyear{impett2016} & \checkmark & \checkmark & \checkmark & $\checkmark^{1,2}$ & & \checkmark \\
Painter by Numbers \cite{painter-by-numbers} & \citeauthor{painter-by-numbers} & \citeyear{painter-by-numbers} & \checkmark & & \checkmark & & \checkmark & \\
Visual Link \cite{visual-link} & \citeauthor{visual-link} & \citeyear{visual-link} & \checkmark & & \checkmark & & & \checkmark \\ 
\peopleart\ \cite{westlake2016} & \citeauthor{westlake2016} & \citeyear{westlake2016} & \checkmark & \checkmark & \checkmark & $\checkmark^{1}$ & \checkmark & \\ 
Art500k \cite{art500k} & \citeauthor{art500k} & \citeyear{art500k} & \checkmark & & \checkmark & & \checkmark & \\ 
BibleVSA \cite{biblevsa} & \citeauthor{biblevsa} & \citeyear{biblevsa} & \checkmark & \checkmark & \checkmark & $\checkmark^{1}$ & & \checkmark \\
ARTigo \cite{artigo} & \citeauthor{artigo} & \citeyear{artigo} & \checkmark & \checkmark & \checkmark & & \checkmark & \\ 
SemArt \cite{semart} & \citeauthor{semart} & \citeyear{semart} & \checkmark & \checkmark & \checkmark & & \checkmark & \\ 
IconArt \cite{iconart} & \citeauthor{iconart} & \citeyear{iconart} & $\color{lightgray}\checkmark$ & \checkmark & \checkmark &  $\checkmark^{1}$ & \checkmark & \\ 
OmniArt \cite{omniart} & \citeauthor{omniart} & \citeyear{omniart} & \checkmark & & \checkmark & & \checkmark & \\ 
MultitaskPainting100k \cite{multitaskpainting100k} & \citeauthor{multitaskpainting100k} & \citeyear{multitaskpainting100k} & \checkmark & & \checkmark & & \checkmark & \\ 
Ancient Chinese Art \cite{sheng2019} & \citeauthor{sheng2019} & \citeyear{sheng2019} & $\color{lightgray}\checkmark$ & \checkmark & \checkmark & & & \checkmark \\
Ancient Egyptian Art \cite{sheng2019} & \citeauthor{sheng2019} & \citeyear{sheng2019} & $\color{lightgray}\checkmark$ & \checkmark & \checkmark & & & \checkmark \\
Artpedia \cite{artpedia} & \citeauthor{artpedia} & \citeyear{artpedia} & \checkmark & \checkmark & \checkmark & & \checkmark & \\ 
Iconclass Caption \cite{cetinic2020} & \citeauthor{cetinic2020} & \citeyear{cetinic2020} & & \checkmark & \checkmark & & & \checkmark \\
AQUA \cite{aqua} & \citeauthor{aqua} & \citeyear{aqua} & $\color{lightgray}\checkmark$ & \checkmark & \checkmark & & \checkmark & \\ 
ClassArch \cite{madhu2020b} & \citeauthor{madhu2020b} & \citetalias{madhu2020b} & \checkmark & \checkmark & \checkmark & $\checkmark^{1,2}$ & & \checkmark \\
Iconclass AI Test Set \cite{iconclass-test-set} & \citeauthor{iconclass-test-set} & \citeyear{iconclass-test-set} & & \checkmark & \checkmark & & \checkmark & \\ 
Saints \cite{schneider2020} & \citeauthor{schneider2020} & \citeyear{schneider2020} & \checkmark & \checkmark & \checkmark & & & \checkmark \\
ArtDL \cite{artdl} & \citeauthor{artdl} & \citeyear{artdl} & & \checkmark & \checkmark & & \checkmark & \\ 
The Met \cite{the-met} & \citeauthor{the-met} & \citeyear{the-met} & \checkmark & & \checkmark & & \checkmark & \\
ArtBench-10 \cite{artbench-10} & \citeauthor{artbench-10} & \citeyear{artbench-10} & \checkmark & & \checkmark & & \checkmark & \\
DEArt \cite{deart} & \citeauthor{deart} & \citeyear{deart} & $\color{lightgray}\checkmark$ & \checkmark & \checkmark & $\checkmark^{1}$ & \checkmark & \\ 
\midrule
\popart\ & Schneider and Vollmer & 2023 & \checkmark & \checkmark & \checkmark & $\checkmark^{1,2}$ & \checkmark & \\
\bottomrule
\end{xltabular}
\raggedright
\textsuperscript{1}Object-level annotations include bounding boxes.\\
\textsuperscript{2}Object-level annotations include keypoints.
\end{table}

\section{Related Work}
\label{chp:related-work}

With the advent of increasingly powerful deep-learning architectures in recent years, the range of domains utilizing computational models has expanded decisively. In the field of Computer Vision, e.g., not only real-world imagery is dealt with anymore, but also figurative representations of imagined phenomena, which are prevalent in art, and across various phases of art history. However, due to those collections' highly original visuals, domain-specific, sufficiently large data sets are still required for training and fine-tuning models.

Prior to the creation of the \popart\ data set, we conducted an extensive study, aggregated in \ref{tab:related-data-sets}, reviewing existing art-historical data sets that can be leveraged for image classification and object recognition tasks. Neither did we consider data sets featuring solely contemporary or born-digital art~\cite{bam}, nor cultural institutions that, while offering relevant data on their websites, do not explicitly make them available in downloadable form, but require prior harvesting.\footnote{For institutions from the \acs{GLAM} (\acl{GLAM}) sector that have published open access data, see the following survey: \url{https://docs.google.com/spreadsheets/d/1WPS-KJptUJ-o8SXtg00llcxq0IKJu8eO6Ege\_GrLaNc/edit}.} We also excluded data sets that are exclusively applicable to other research areas like aesthetic quality assessment~\cite{jenaesthetics}, sentiment analysis~\cite{mart, wikiart-emotions}, or correspondence matching~\cite{brueghel, jenicek2019}. While formal attributes at the image-level are contained in a large number of data sets, enabling the classification of artists, materials, or creation dates, among others~\cite{wikiart, painting-91, rijksmuseum-challenge, painter-by-numbers, art500k, omniart, multitaskpainting100k, artpedia, the-met, artbench-10}, content-based tags are less frequent. This is due to the fact that labels referring to the image phenomena actually shown must be determined by manual annotation, driven either by crowdsourcing approaches~\cite{artigo} or singular institutional efforts~\cite{iconart, cetinic2020, iconclass-test-set}. The latter rely on the iconographic classification system Iconclass, which is conceived for the Western motifs of the visual arts~\cite{vandeWaal1973}. As a result of the already time-consuming labeling process at image-level, few data sets feature object-level annotations~\cite{yarlagadda2010, printart, impett2016, westlake2016, biblevsa, iconart, madhu2020b, deart}. When provided, they are usually marked with bounding boxes, so that object instances are enclosed with rectangles and thus precisely located in the image. To our work here of particular importance is the \peopleart\ data set~\cite{westlake2016}, in which human figures shown in nearly $1{,}500$ images are labeled with bounding boxes. Unlike the ten times larger \deart\ data set~\cite{deart}, which identifies figures in collections only from the \nth{12} to \nth{18} centuries, \peopleart\ indicates depiction styles that encompass Impressionist movements as well as Surrealist ones with rather artificial forms of body representation.

For the decoding of human poses, the rectangular framing of the entire body is not sufficient: individual limbs cannot be identified and differentiated any more than joints, such as elbows and wrists. To obtain more accurate information about the position of articulation points, three annotation practices have been used. \citet{deart} roughly classify poses into $12$~categories, e.g., by labeling human figures as sitting or kneeling. \citet{printart}, on the other hand, place additional bounding boxes around the torso and head to approximate the specifics of the human body. Only \citet{impett2016} and \citet{madhu2020b}, however, apply fine-grained labels to faithfully represent bodily specifics by assigning keypoints on areas relevant to the figure's pose, e.g., the hips, knees, or ears. In doing so, they adhere to labeling techniques common for real-world human pose estimation. The Microsoft \ac{COCO} format guidelines, for instance, require that $17$~keypoints be stored with their $xy$-coordinates.\footnote{\url{https://cocodataset.org/\#format-data}.} Both data sets suffer from two issues: they are (i)~not made publicly available for further reuse, and (ii)~devoted to only a comparatively narrow subset of art-historical modes of depicting human figures; \citet{impett2016} extracted panels from Warburg's Bilderatlas \textit{Mnemosyne}, whereas \citet{madhu2020b} focused on ancient Greek vase paintings. With \popart, we address this desideratum and introduce the first publicly available data set for human pose estimation in art-historical figures, covering impressionistic to neo-figurative and realistic depiction styles. Since our data set follows the Microsoft \ac{COCO} format~\cite{lin2014}, in addition to bounding boxes, up to $17$~keypoints are stored per figure. 
Five keypoints are provided for the head, indicating the nose, eyes, and ears; six for the upper body, indicating wrists, elbows, and shoulders; and another six for the lower body, indicating ankles, knees, and hips.

\section{Data Set}
\label{chp:data-set}

This section elaborates on the acquisition and constitution of the \popart\ data set. First, we outline the image collection (Section~\ref{chp:image-collection}) and annotation procedures (Section~\ref{chp:image-annotation}). We then provide an in-depth statistical analysis of the data set (Section~\ref{chp:descriptive-statistics}) and present its underlying data format (Section~\ref{chp:data-split-format}).

\begin{table}[t!]
\footnotesize
\begin{xltabular}{\columnwidth}{@{}XlT*{7}{S}@{}}
\caption{
Figure detection results are reported for the \peopleart\ test set~\cite{westlake2016}. For training and validation, \peopleart\ is used as well. In contrast to previous benchmarks by \citet{kadish2021} and \citet{gonthier2022}, we include difficult-to-annotate figures. The best performing approach is indicated in bold.
}
\label{tab:detection-benchmark-people-art}\\
\toprule
Model & Backbone & LR & $\text{AP}$ & $\text{AP}_{50}$ & $\text{AP}_{75}$ & $\text{AP}_{S}$ & $\text{AP}_{M}$ & $\text{AP}_{L}$ & $\text{AR}$\\
\midrule
\ac{TOOD} \cite{tood} & ResNet-50-FPN & $2\textrm{e}-4$ & 0.461 & 0.750 & \textbf{0.490} & \textbf{0.197} & 0.296 & 0.493 & \textbf{0.635}\\
\ac{PVT} \cite{pvt} & PVTv2-B2 & $1\textrm{e}-5$ & \textbf{0.465} & \textbf{0.760} & 0.484 & 0.060 & 0.263 & \textbf{0.505} & 0.601\\
Cascade R-CNN \cite{cascade_rcnn} & ResNet-50-FPN & $2\textrm{e}-4$ & 0.444 & 0.758 & 0.468 & 0.147 & 0.297 & 0.476 & 0.593\\
SABL Cascade R-CNN \cite{sabl_faster_rcnn} & ResNet-50-FPN & $2\textrm{e}-4$ & 0.443 & 0.741 & 0.458 & 0.139 & 0.286 & 0.476 & 0.593\\
Faster R-CNN \cite{faster_rcnn} & ResNet-50-FPN & $2\textrm{e}-4$ & 0.423 & 0.749 & 0.421 & 0.115 & \textbf{0.298} & 0.450 & 0.568\\
SABL Faster R-CNN \cite{sabl_faster_rcnn} & ResNet-50-FPN & $2\textrm{e}-4$ & 0.441 & 0.752 & 0.466 & 0.123 & 0.284 & 0.475 & 0.596\\
PISA Faster R-CNN \cite{pisa_faster_rcnn} & ResNet-50-FPN & $2\textrm{e}-4$ & 0.434 & 0.753 & 0.451 & 0.137 & 0.290 & 0.463 & 0.568\\
Libra Faster R-CNN \cite{libra_faster_rcnn} & ResNet-50-FPN & $2\textrm{e}-4$ & 0.417 & 0.747 & 0.416 & 0.068 & 0.290 & 0.445 & 0.569\\
\bottomrule
\end{xltabular}
\end{table}

\subsection{Image Collection}
\label{chp:image-collection}

Like many authors before, e.g., \citet{westlake2016} and \citet{art500k}, we exploit the art-historical online encyclopedia \wikiart\ \cite{wikiart} as content provider. This decision is attributable to several factors: (i)~reproductions provided in \wikiart\ are mostly in the public domain and can thus be redistributed under free licenses; (ii)~not only does \wikiart\ embrace the widely received canon of Western art history, but does also include Eastern movements, such as the early \nth{20}-century Japanese Shin-hanga, albeit to a much lesser extent; (iii)~because \wikiart\ stores the depiction style of each object, fine-grained evaluations are facilitated, even if such classifications are to be understood as loose, arbitrary, or possibly biased constructs~\cite{camille1996, elkins2007}.

\begin{figure}
\centering
\begin{subfigure}{.24\linewidth}
  \centering
  \includegraphics[width=\linewidth]{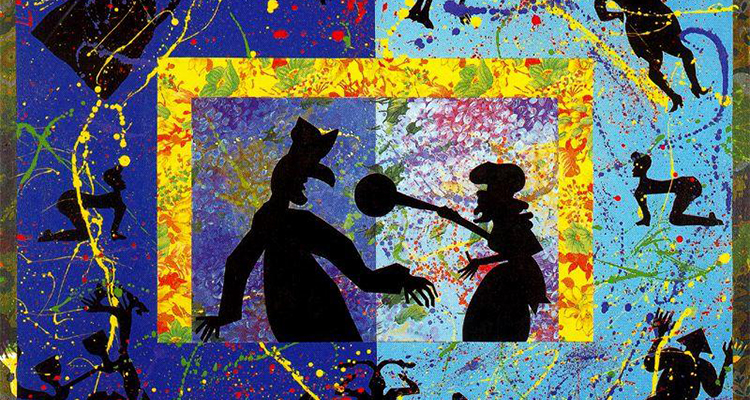}
  \caption{Abstract Expressionism\vspace{5pt}}
\end{subfigure}%
\hfill
\begin{subfigure}{.24\linewidth}
  \centering
  \includegraphics[width=\linewidth]{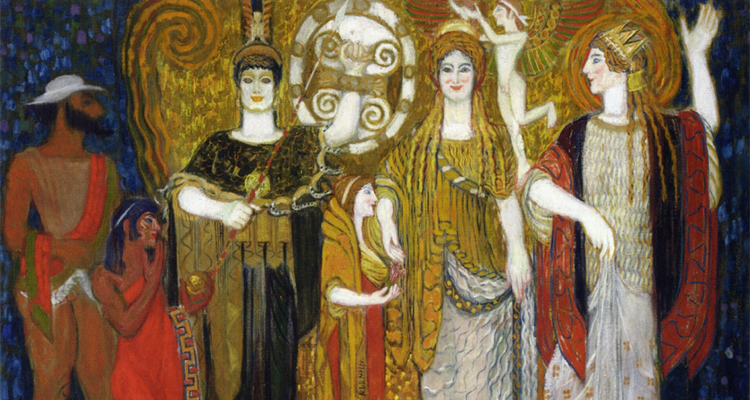}
  \caption{Art Nouveau\vspace{5pt}}
\end{subfigure}%
\hfill
\begin{subfigure}{.24\linewidth}
  \centering
  \includegraphics[width=\linewidth]{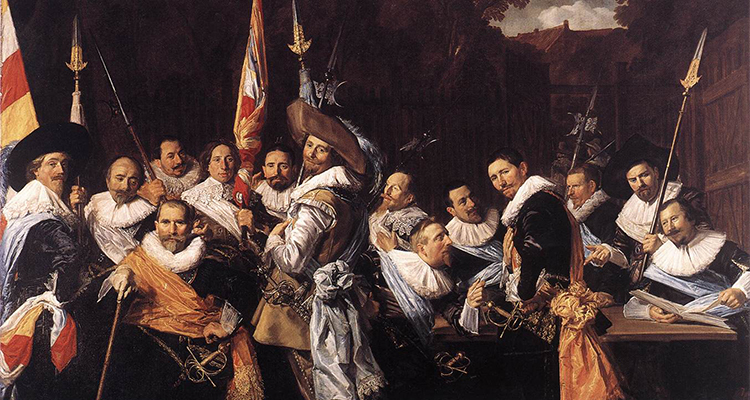}
  \caption{Baroque\vspace{5pt}}
\end{subfigure}%
\hfill
\begin{subfigure}{.24\linewidth}
  \centering
  \includegraphics[width=\linewidth]{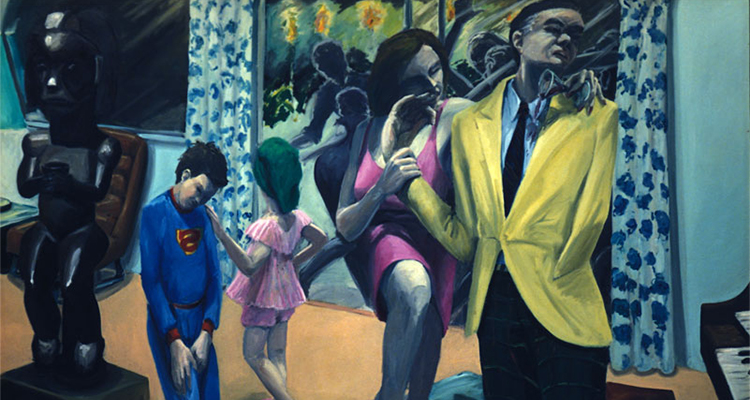}
  \caption{Contemporary Realism\vspace{5pt}}
\end{subfigure}%

\begin{subfigure}{.24\linewidth}
  \centering
  \includegraphics[width=\linewidth]{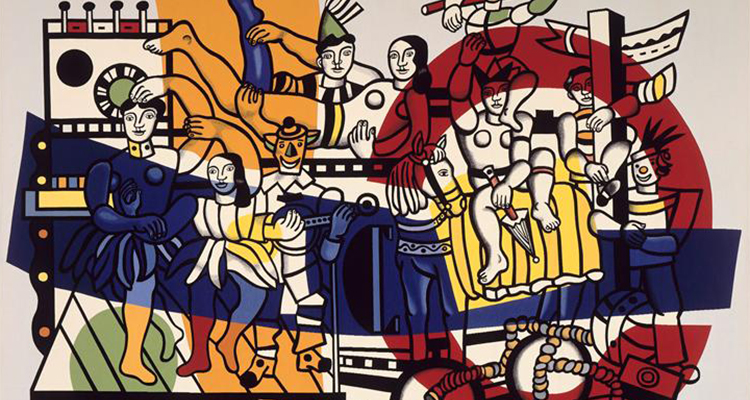}
  \caption{Cubism\vspace{5pt}}
\end{subfigure}%
\hfill
\begin{subfigure}{.24\linewidth}
  \centering
  \includegraphics[width=\linewidth]{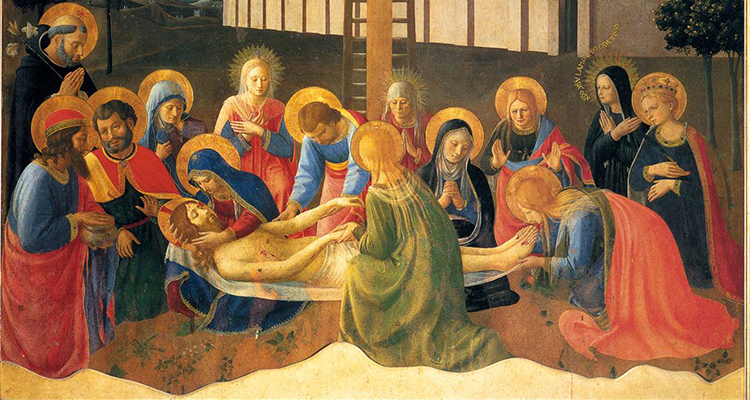}
  \caption{Early Renaissance\vspace{5pt}}
\end{subfigure}%
\hfill
\begin{subfigure}{.24\linewidth}
  \centering
  \includegraphics[width=\linewidth]{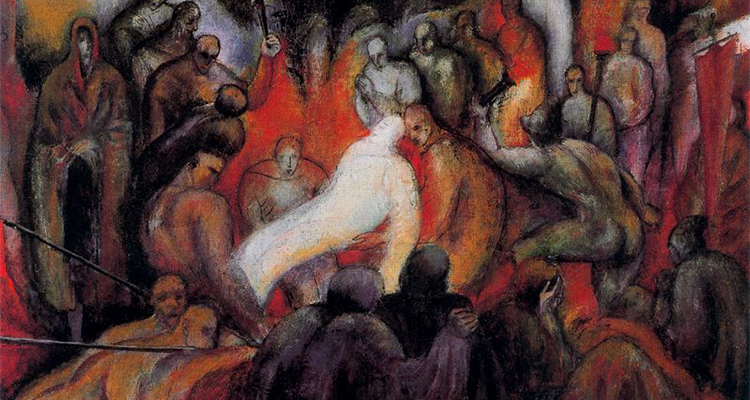}
  \caption{Expressionism\vspace{5pt}}
\end{subfigure}%
\hfill
\begin{subfigure}{.24\linewidth}
  \centering
  \includegraphics[width=\linewidth]{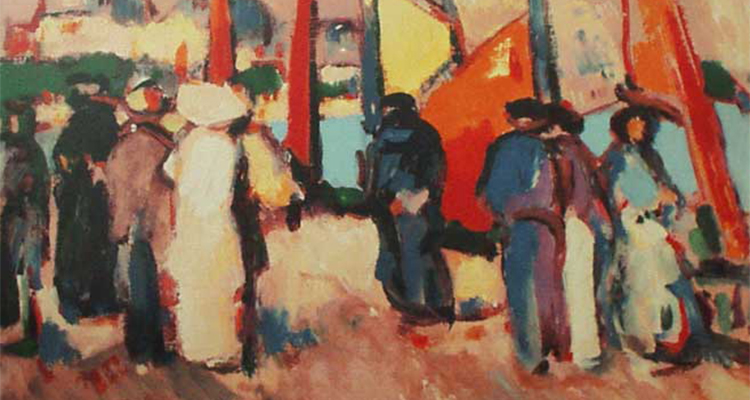}
  \caption{Fauvism\vspace{5pt}}
\end{subfigure}%

\begin{subfigure}{.24\linewidth}
  \centering
  \includegraphics[width=\linewidth]{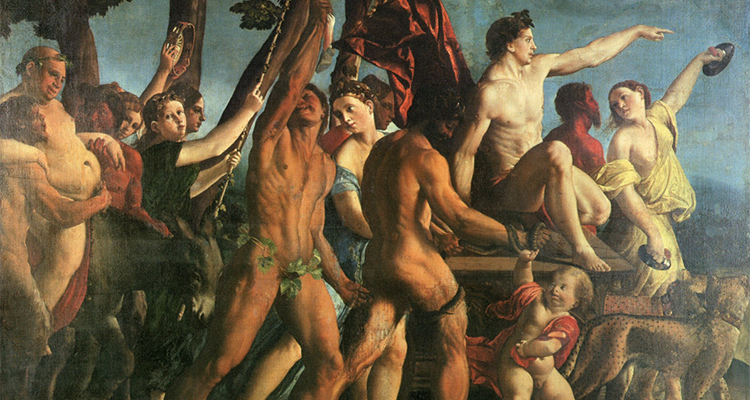}
  \caption{High Renaissance\vspace{5pt}}
\end{subfigure}%
\hfill
\begin{subfigure}{.24\linewidth}
  \centering
  \includegraphics[width=\linewidth]{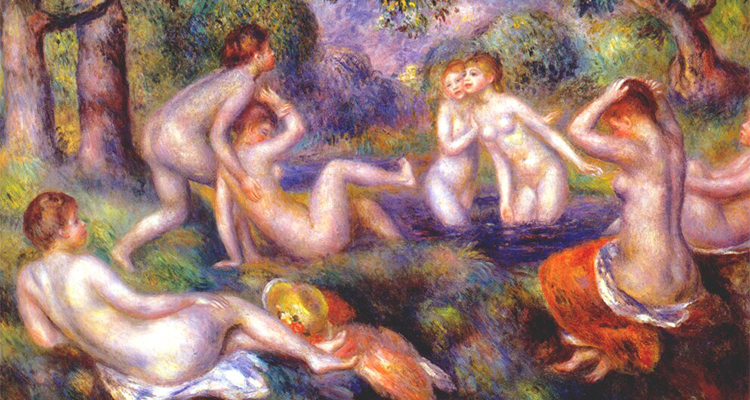}
  \caption{Impressionism\vspace{5pt}}
\end{subfigure}%
\hfill
\begin{subfigure}{.24\linewidth}
  \centering
  \includegraphics[width=\linewidth]{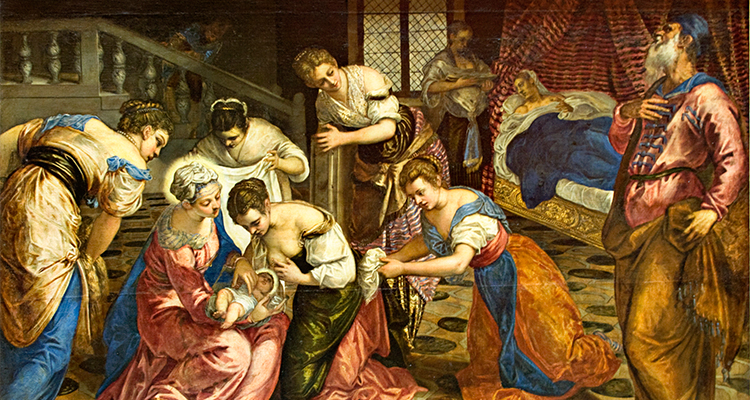}
  \caption{Mannerism\vspace{5pt}}
\end{subfigure}%
\hfill
\begin{subfigure}{.24\linewidth}
  \centering
  \includegraphics[width=\linewidth]{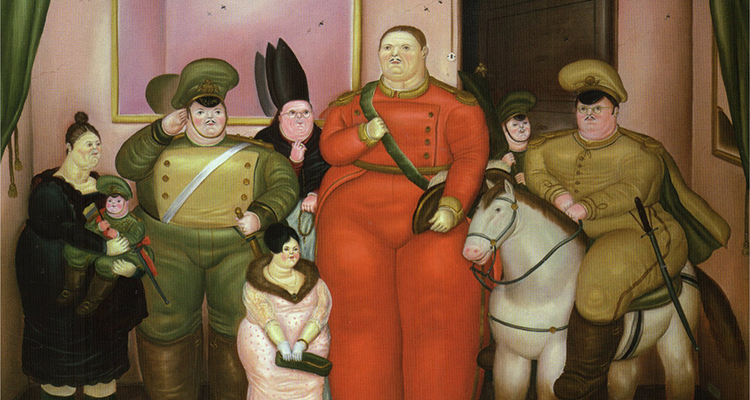}
  \caption{Naive Art\vspace{5pt}}
\end{subfigure}%

\begin{subfigure}{.24\linewidth}
  \centering
  \includegraphics[width=\linewidth]{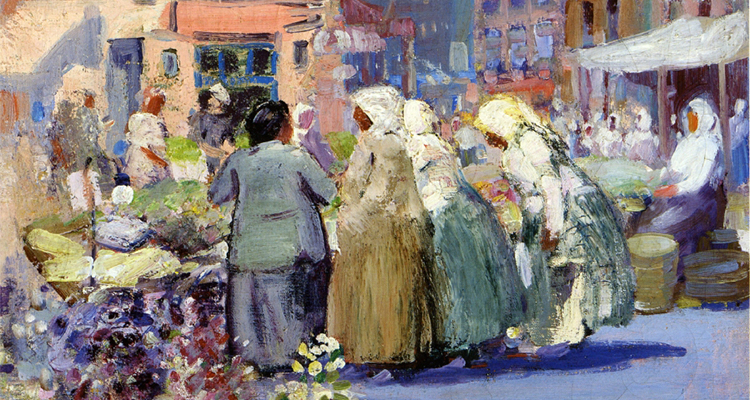}
  \caption{New Realism\vspace{5pt}}
\end{subfigure}%
\hfill
\begin{subfigure}{.24\linewidth}
  \centering
  \includegraphics[width=\linewidth]{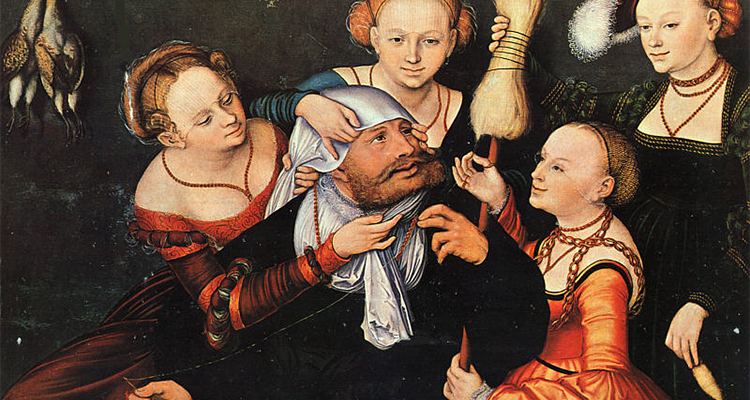}
  \caption{Northern Renaissance\vspace{5pt}}
\end{subfigure}%
\hfill
\begin{subfigure}{.24\linewidth}
  \centering
  \includegraphics[width=\linewidth]{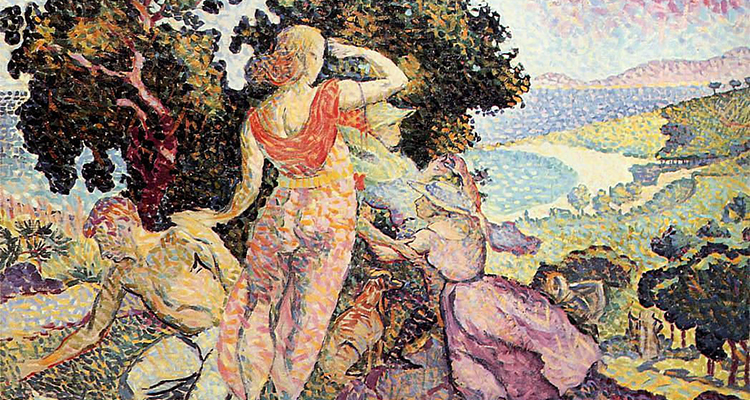}
  \caption{Pointillism\vspace{5pt}}
\end{subfigure}%
\hfill
\begin{subfigure}{.24\linewidth}
  \centering
  \includegraphics[width=\linewidth]{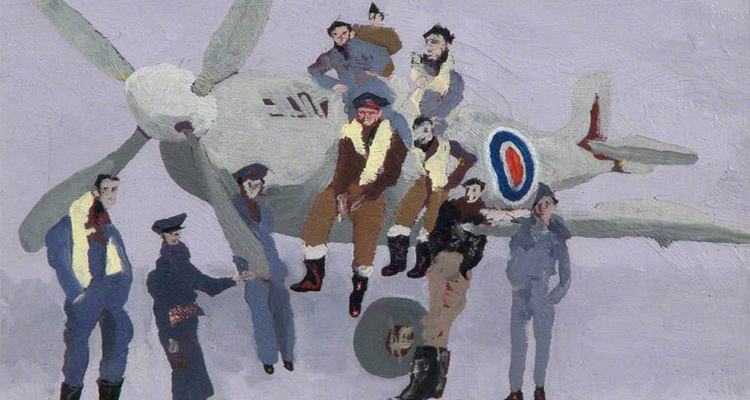}
  \caption{Pop Art\vspace{5pt}}
\end{subfigure}%

\begin{subfigure}{.24\linewidth}
  \centering
  \includegraphics[width=\linewidth]{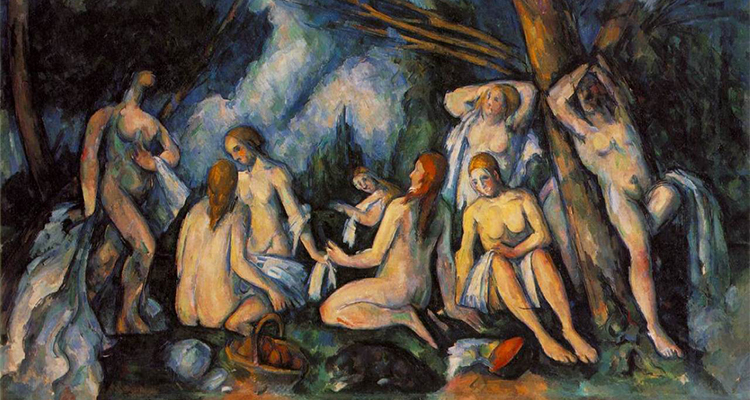}
  \caption{Post Impressionism\vspace{5pt}}
\end{subfigure}%
\hfill
\begin{subfigure}{.24\linewidth}
  \centering
  \includegraphics[width=\linewidth]{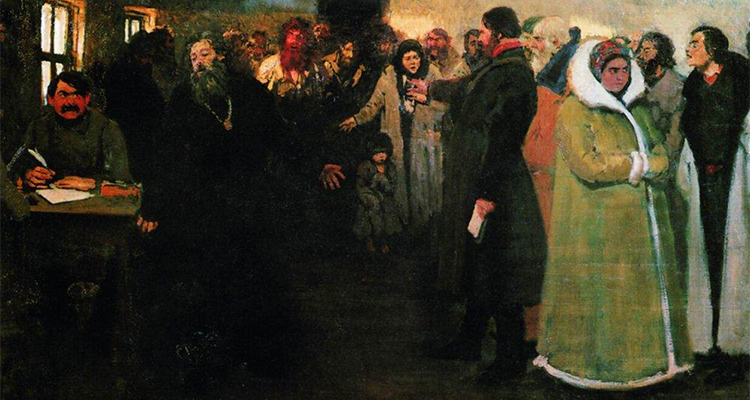}
  \caption{Realism\vspace{5pt}}
\end{subfigure}%
\hfill
\begin{subfigure}{.24\linewidth}
  \centering
  \includegraphics[width=\linewidth]{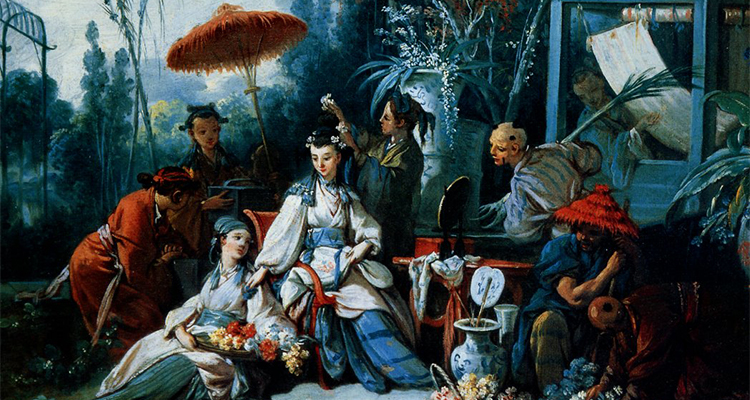}
  \caption{Rococo\vspace{5pt}}
\end{subfigure}%
\hfill
\begin{subfigure}{.24\linewidth}
  \centering
  \includegraphics[width=\linewidth]{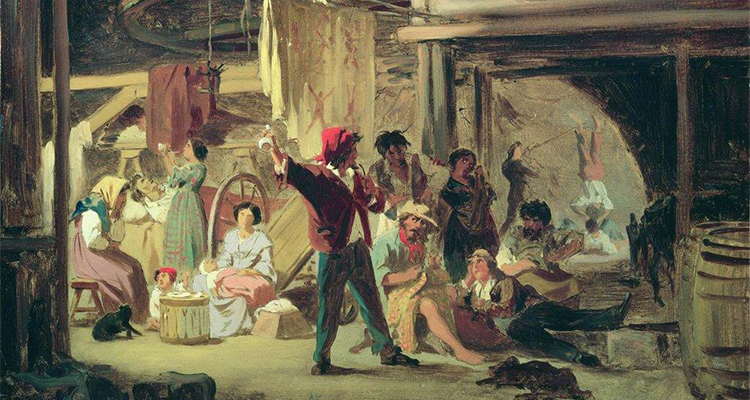}
  \caption{Romanticism\vspace{5pt}}
\end{subfigure}%

\begin{subfigure}{.24\linewidth}
  \centering
  \includegraphics[width=\linewidth]{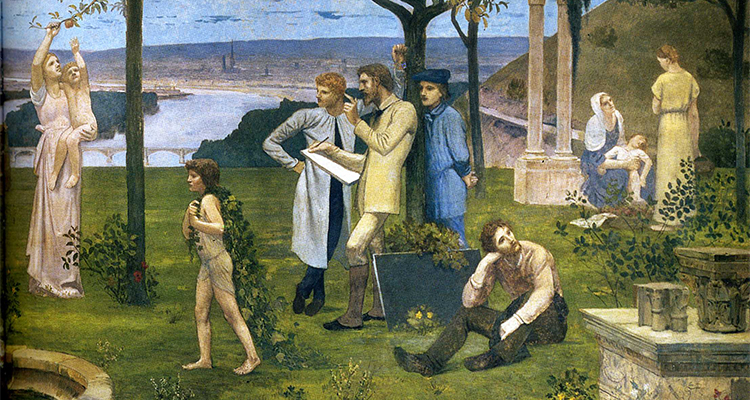}
  \caption{Symbolism}
\end{subfigure}%
\hfill
\begin{subfigure}{.24\linewidth}
  \centering
  \includegraphics[width=\linewidth]{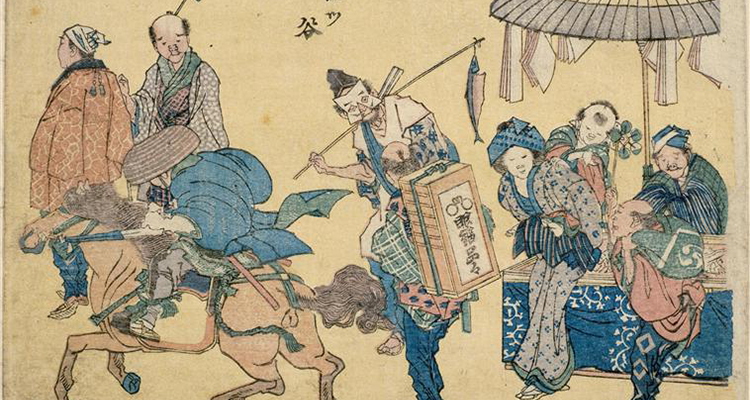}
  \caption{Ukiyo-e}
\end{subfigure}%
\hfill
\begin{subfigure}{.24\linewidth}
\phantom{Test}
\end{subfigure}%
\hfill
\begin{subfigure}{.24\linewidth}
\phantom{Test}
\end{subfigure}%
\caption{
The \popart\ data set contains $22$ depiction styles, ranging from impressionistic to neo-figurative and realistic variants. For each style, an exemplary image is shown. All images originate from the art-historical online encyclopedia \wikiart~\cite{wikiart} and are in the public domain.
}
\label{fig:style-overview-popart}
\end{figure}

To further ensure that \popart\ is representative of both the projective and denotational styles prevalent in the domain~\cite{willats1997}, a semi-automatic data collection procedure was preferred. In a preliminary step, we extracted images from \wikiart\ that have a high probability of depicting human figures, i.e., images on which at least one figure can be automatically detected with a probability of $p=0.5$. To this end, we benchmarked the suitability of models commonly used for object detection and applied the best-performing one. The selection ranges from multi-stage \acp{R-CNN}~\cite{faster_rcnn, pisa_faster_rcnn, libra_faster_rcnn, sabl_faster_rcnn, cascade_rcnn} and Transformer-based architectures~\cite{pvt} to task-aligned one-stage methods~\cite{tood}. For evaluation, we use the metrics and tools provided by the \ac{COCO} API.\footnote{\url{https://github.com/cocodataset/cocoapi}.} All models were first pre-trained on the Microsoft \ac{COCO} 2017 data set\footnote{\url{https://www.kaggle.com/datasets/awsaf49/coco-2017-dataset}.} for $12$ epochs. As optimization algorithms, we employed \ac{SGD} for ResNet-50 and Adam~\cite{adam} for Transformer backbones; momentum and weight decay were set to $0.9$ and $1\textrm{e}-4$, respectively. The initial learning rate decays at the \nth{8} and \nth{11} epoch with $2\textrm{e}-2$ set for ResNet-50-backed and $1\textrm{e}-4$ for Transformer-backed architectures. Models were then fine-tuned, with their classification head re-initialized, for another $12$ epochs on \peopleart~\cite{westlake2016}. The learning rate is decreased to $2\textrm{e}-4$ in case of ResNet-50 and $1\textrm{e}-5$ in case of Transformer backbones. During training, we adopted the following data augmentation techniques from the Albumentations library~\cite{albumentations} to increase the models' robustness: (i)~either \texttt{RandomBrightnessContrast} or \texttt{CLAHE} is applied with a probability of $p=0.2$; (ii)~either \texttt{RGBShift} or \texttt{HueSaturationValue} is applied with $p=0.1$; (iii)~\texttt{JpegCompression} is applied with $p=0.2$; (iv)~\texttt{ChannelShuffle} is applied with $p=0.1$; and (v)~either \texttt{Blur} or \texttt{MedianBlur} is applied with $p=0.1$. Images are reduced to a maximum scale of $1,333 \times 800$ pixels without changing the aspect ratio. In contrast to previous studies by \citet{kadish2021} and \citet{gonthier2022}, we include difficult-to-annotate figures. As evident by the benchmark results shown in \ref{tab:detection-benchmark-people-art}, state-of-the-art models such as \ac{TOOD}~\cite{tood} and \ac{PVT}~\cite{pvt} outperform multistage \acp{R-CNN} to a nearly similar extent in \ac{AP} between $1.7$ and $4.8$\,\%. At a more restrictive \ac{IoU} threshold of $0.75$, the difference increases further, rising to between $1.6$ and $7.4$\,\%. This effect also is noticeable with \ac{AR}, which is $0.5$ to $6.8$\,\% higher. Since \ac{TOOD} surpasses \ac{PVT} in \ac{AR} by $3.4$\,\%, with \ac{AP} being almost equal, we assume that it is generally suited best to the stylistic peculiarities of the art-historical domain.

After pre-filtering the data for images with human figures, we identified the $22$~most frequently observed depiction styles, covering impressionistic, neo-figurative, and realistic movements from the \nth{14} to the \nth{20} century. The integration of data from the \nth{19} and \nth{20} centuries is of particular importance here, as formal conventions of bodily phenomena were successively disrupted 
at the end of the \nth{19} century~\cite{butterfield2021}. We deemed $22$~styles to be adequate to both capture the wide diversity of art-historical image specifics in a time-efficient manner, and to later sufficiently assess the validity of computational models for bounding box and keypoint estimation depending on the depiction style. A maximum of $125$~images per style were then selected for image annotation, taking into account the sampling distribution. Exact-duplicate and near-duplicate reproductions were removed. For each style, an example image is shown in \ref{fig:style-overview-popart}.

\begin{figure}
\centering
\begin{subfigure}{.49\linewidth}
  \centering
  \includegraphics[width=\linewidth]{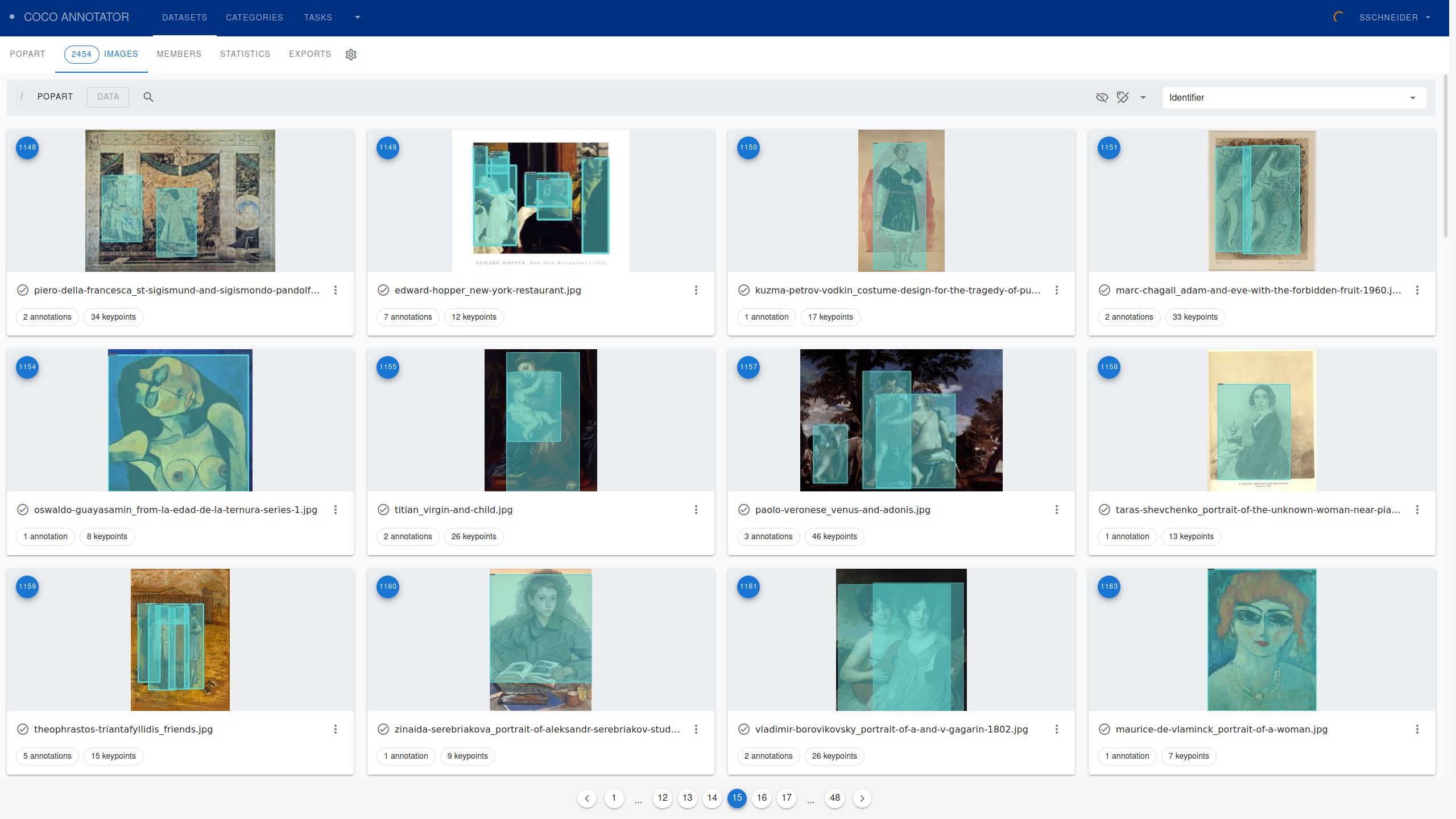}
  \caption{Data set view}
  \label{}
\end{subfigure}%
\hfill
\begin{subfigure}{.49\linewidth}
  \centering
  \includegraphics[width=\linewidth]{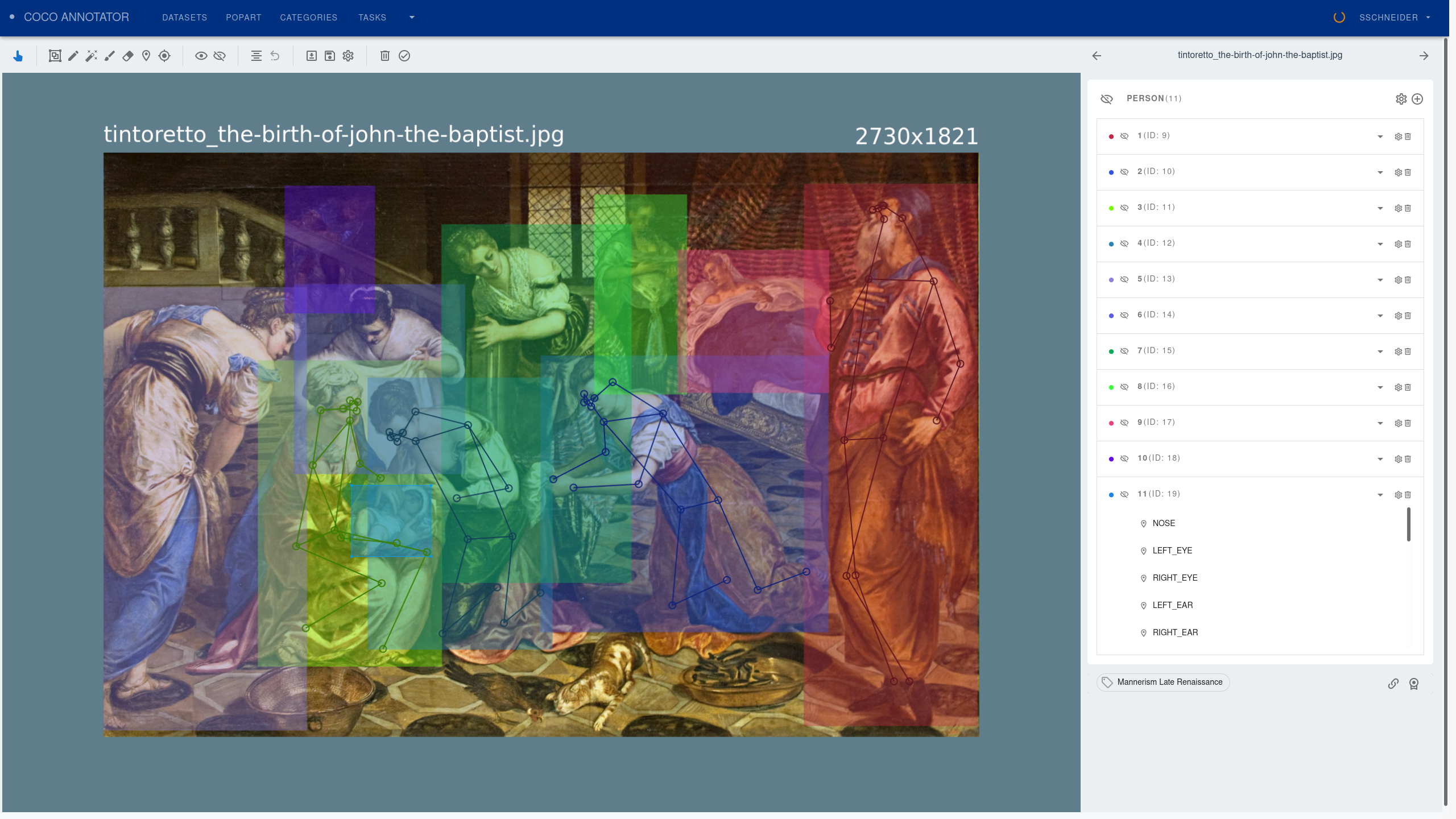}
  \caption{Annotation view}
  \label{}
\end{subfigure}%
\caption{
The web-based open-source tool \ac{COCO} Annotator~\cite{coco-annotator} provides a light-weight interface that can be used collaboratively for annotating bounding boxes and keypoints.
}
\label{fig:coco-annotator}
\end{figure}

\subsection{Image Annotation}
\label{chp:image-annotation}

The practice of image annotation is characterized by two modes of determinations: whether a human figure can be recognized in an image (\textit{bounding box annotation}) and how his or her pose can be abstracted in it 
(\textit{keypoint annotation}). Following \citet{everingham2010}, we designed the annotation procedure to be as (i)~exhaustive, (ii)~consistent, and (iii)~accurate as possible, without omitting art-historical depiction specifics. With \ac{COCO} Annotator~\cite{coco-annotator}, we used a web-based open-source tool for bounding box and keypoint annotation that we minimally adapted to our needs (\ref{fig:coco-annotator}). 

\begin{figure}
\scriptsize

\begin{forest}
  for tree={
    forked edges,
    l sep+=2.5pt,
    for tree={
      grow'=0,
      align=left,
      anchor=west,
      text width=2.75cm,
    },
    where level=0{text width=1.25cm}{},
  }
  [Challenges
    [Variations of the size \\of human figures
      [Large crowds, text width=2.55cm, name=A_0]
      [\phantom{Test}, text width=2.55cm, l=0.875cm, no edge, tikz={\draw (.east)|-(A_0.east);\draw (.east)|-(B_0.east);}
        [Small figures]
        [Figures hard to separate \\from each other]
        [Figures difficult to \\recognize as such
          [Image-extrinsic factors, edge=densely dotted]
          [Image-intrinsic factors, edge=densely dotted]
        ]
      ]
      [Figures in the \\background, text width=2.55cm, name=B_0 
      ]
    ]
    [Relation of human \\figures to each other
      [Referencing
        [Shadows]
        [Reflections
          [In water]
          [In the mirror]
          [On other surfaces \\(such as helmets)]
        ]
      ]
      [Not referencing
        [Overlaps]
        [Intersections]
        [Symmetrically \\arranged figures]
      ]
    ]
    [Deviations from the \\\enquote*{ideal} human body
      [Stylistic variance
        [Lack of differentiation \\of the face]
        [Lack of differentiation \\of the body shape]
        [Veiled body]
      ]
      [Non-human bodies \\and body parts
        [Human-like animals 
        ]
        [Mythological figures 
        ]
        [Biblical figures 
        ]
      ]
      [Non-living bodies \\and body parts
        [Fabricated bodies
          [Dolls]
          [Masks]
          [Crafts]
          [Sculptures] 
        ]
        [Skeletons] 
        [Severed heads]
        [Severed limbs 
          [Image-extrinsic factors, edge=densely dotted]
          [Image-intrinsic factors, edge=densely dotted]
        ]
      ]
    ]
    [Positioning of the \\human body
      [Back views]
      [Profile views
      ]
      [Distortions]
      [Twists and turns]
    ]
  ]
\end{forest}
  
\caption{
Four aspects pose challenges to the annotation of art-historical imagery: (i)~the size of human figures, (ii)~their relation to each other, (iii)~deviations from the \enquote*{ideal} human body, and (iv)~the positioning of the body.
}
\label{fig:image-annotation-taxonomy}
\end{figure}
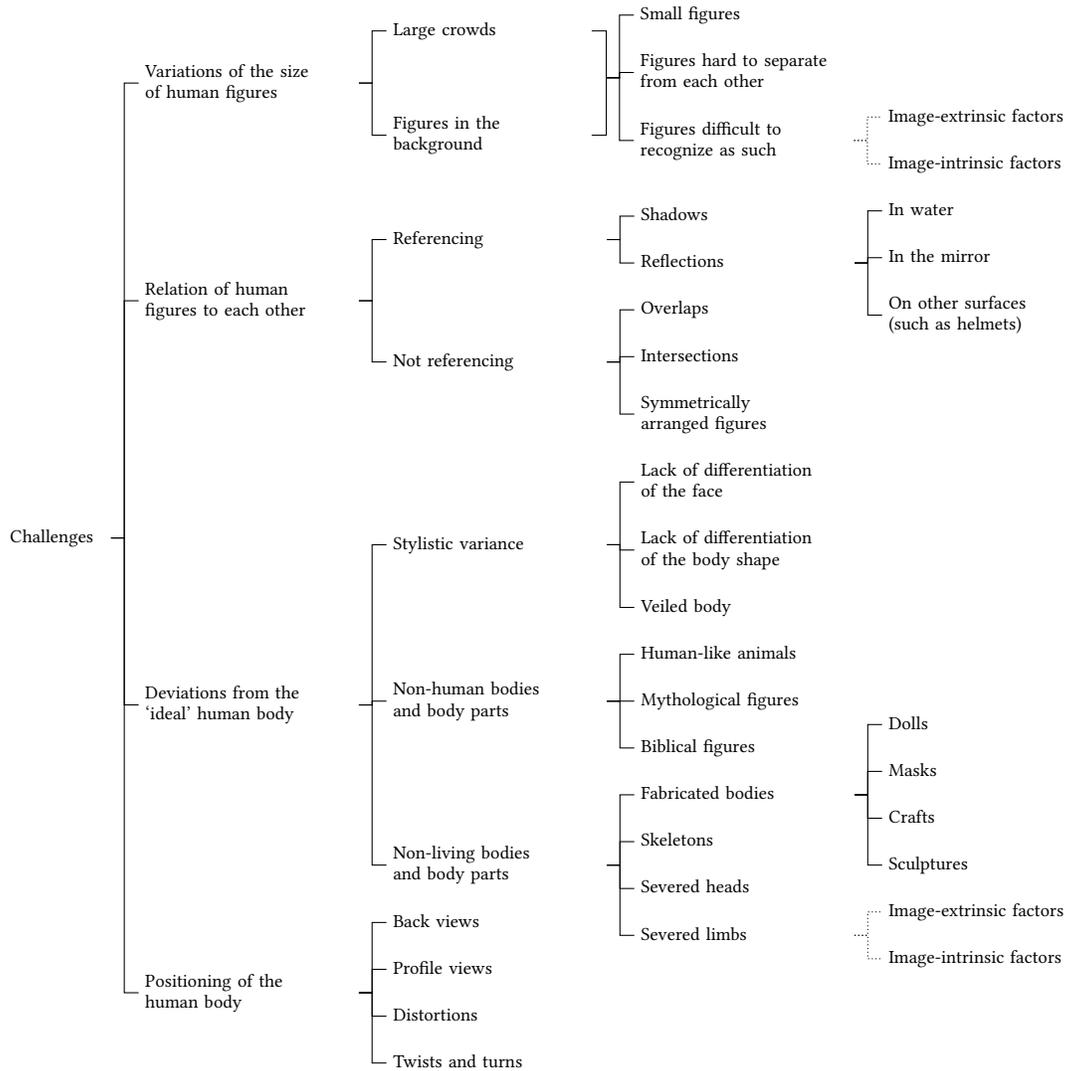
\begin{figure}
\centering

\begin{subfigure}[t]{.05\linewidth}
  \centering
  \rotatebox{90}{\hspace{-22pt}\parbox{2.25cm}{\scriptsize\centering Variations of the size of human figures}}
\end{subfigure}%
\hfill
\begin{subfigure}{.225\linewidth}
  \centering
  \includegraphics[width=\linewidth]{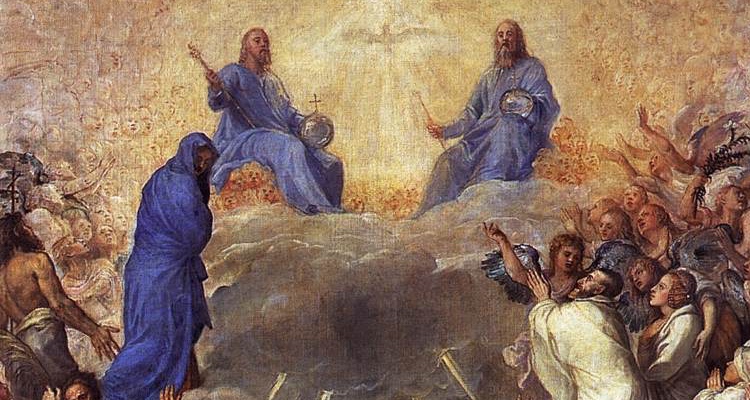}
  \caption{}
  \label{fig:image-annotation-example-1a}
\end{subfigure}%
\hfill
\begin{subfigure}{.225\linewidth}
  \centering
  \includegraphics[width=\linewidth]{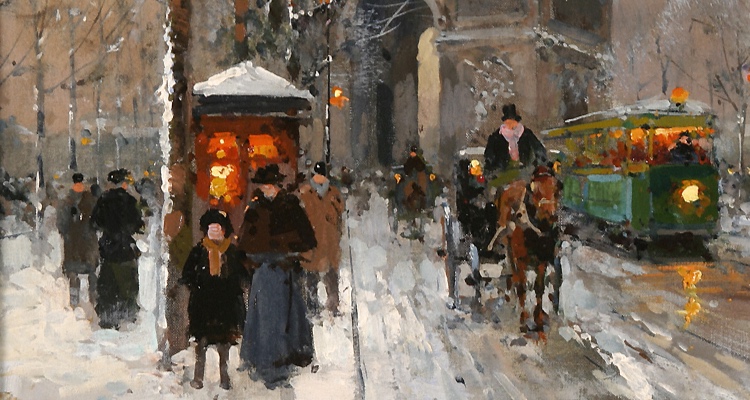}
  \caption{}
  \label{fig:image-annotation-example-1b}
\end{subfigure}%
\hfill
\begin{subfigure}{.225\linewidth}
  \centering
  \includegraphics[width=\linewidth]{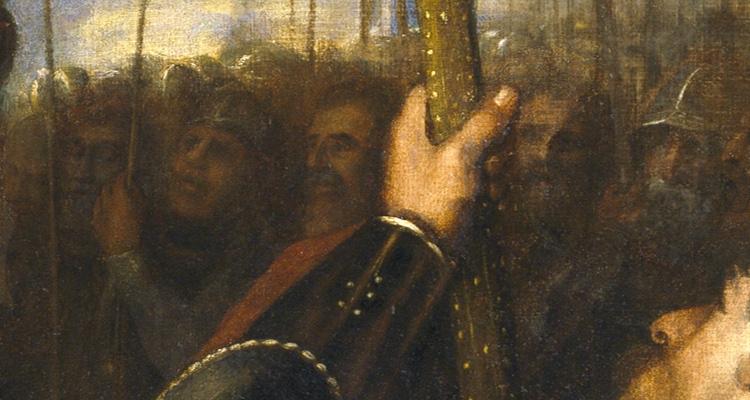}
  \caption{}
  \label{fig:image-annotation-example-1c}
\end{subfigure}%
\hfill
\begin{subfigure}{.225\linewidth}
  \centering
  \includegraphics[width=\linewidth]{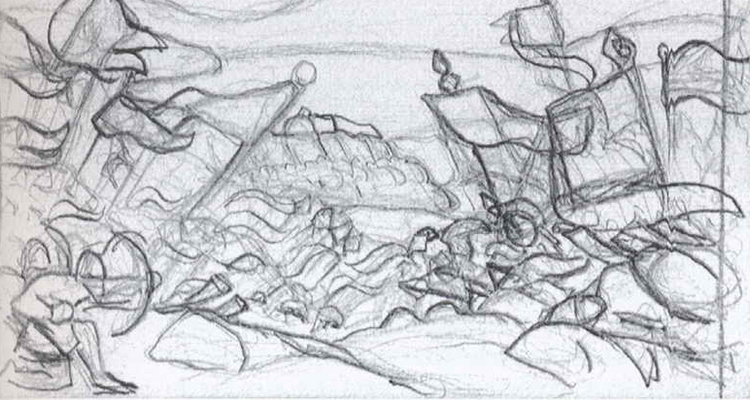}
  \caption{}
  \label{fig:image-annotation-example-1d}
\end{subfigure}%


\begin{subfigure}[t]{.05\linewidth}
  \centering
  \rotatebox{90}{\hspace{-22pt}\parbox{2.25cm}{\scriptsize\centering Relation of human figures to each other}}
\end{subfigure}%
\hfill
\begin{subfigure}{.225\linewidth}
  \centering
  \includegraphics[width=\linewidth]{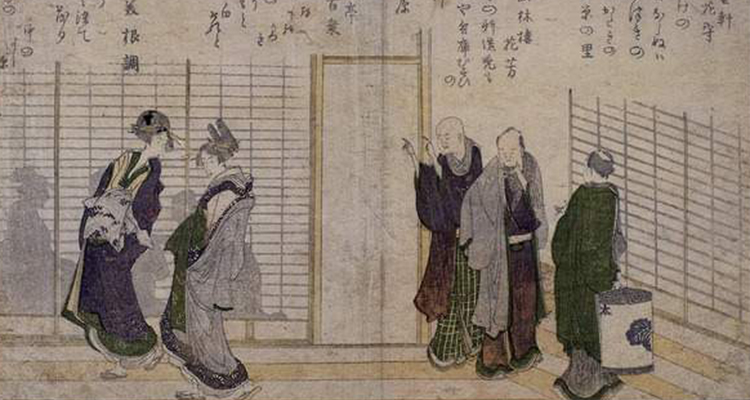}
  \caption{}
  \label{fig:image-annotation-example-2a}
\end{subfigure}%
\hfill
\begin{subfigure}{.225\linewidth}
  \centering
  \includegraphics[width=\linewidth]{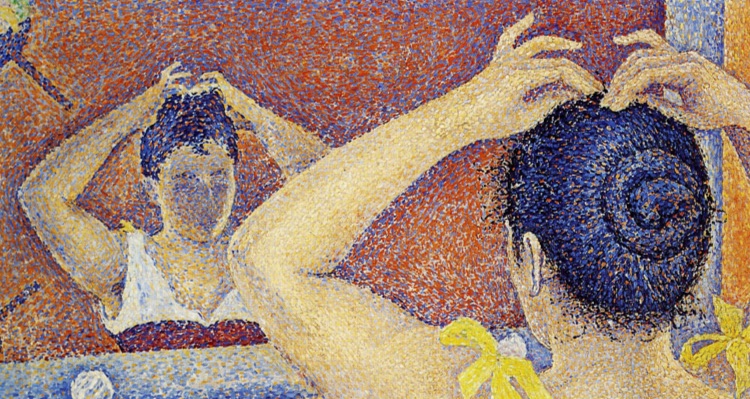}
  \caption{}
  \label{fig:image-annotation-example-2b}
\end{subfigure}%
\hfill
\begin{subfigure}{.225\linewidth}
  \centering
  \includegraphics[width=\linewidth]{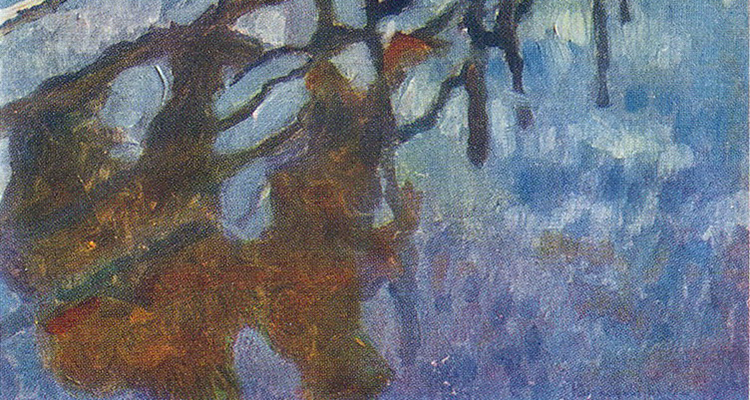}
  \caption{}
  \label{fig:image-annotation-example-2c}
\end{subfigure}%
\hfill
\begin{subfigure}{.225\linewidth}
  \centering
  \includegraphics[width=\linewidth]{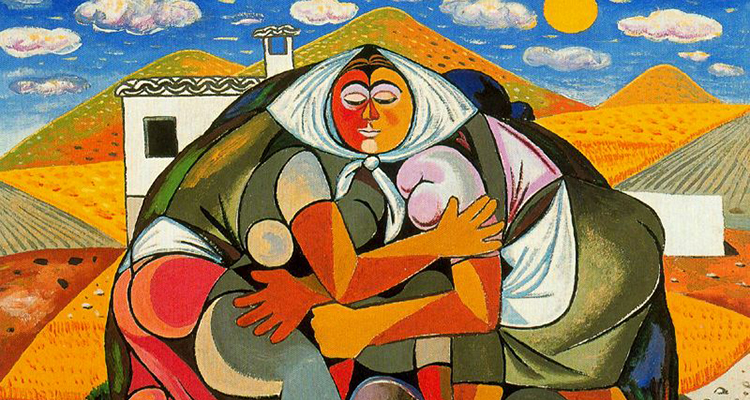}
  \caption{}
  \label{fig:image-annotation-example-2d}
\end{subfigure}%


\begin{subfigure}[t]{.05\linewidth}
  \centering
  \rotatebox{90}{\hspace{-22pt}\parbox{2.25cm}{\scriptsize\centering Deviation from the \enquote*{ideal} human body}}
\end{subfigure}%
\hfill
\begin{subfigure}{.225\linewidth}
  \centering
  \includegraphics[width=\linewidth]{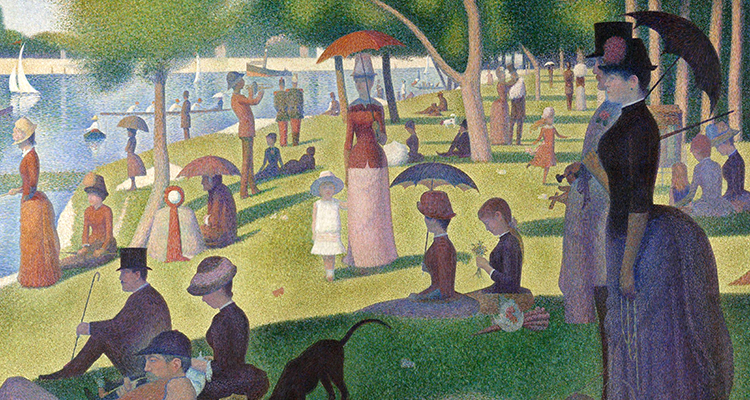}
  \caption{}
  \label{fig:image-annotation-example-3a}
\end{subfigure}%
\hfill
\begin{subfigure}{.225\linewidth}
  \centering
  \includegraphics[width=\linewidth]{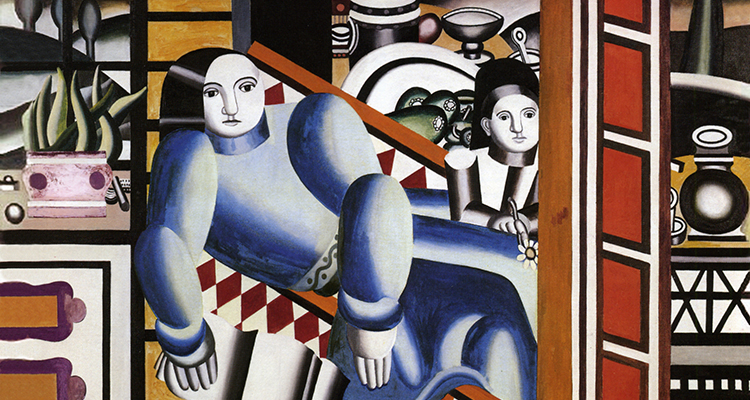}
  \caption{}
  \label{fig:image-annotation-example-3b}
\end{subfigure}%
\hfill
\begin{subfigure}{.225\linewidth}
  \centering
  \includegraphics[width=\linewidth]{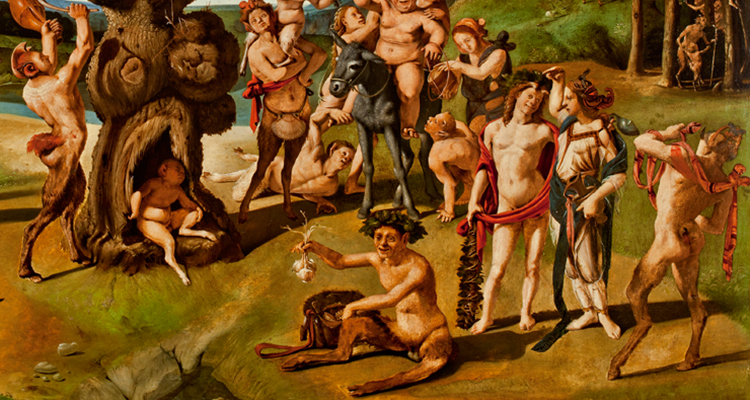}
  \caption{}
  \label{fig:image-annotation-example-3c}
\end{subfigure}%
\hfill
\begin{subfigure}{.225\linewidth}
  \centering
  \includegraphics[width=\linewidth]{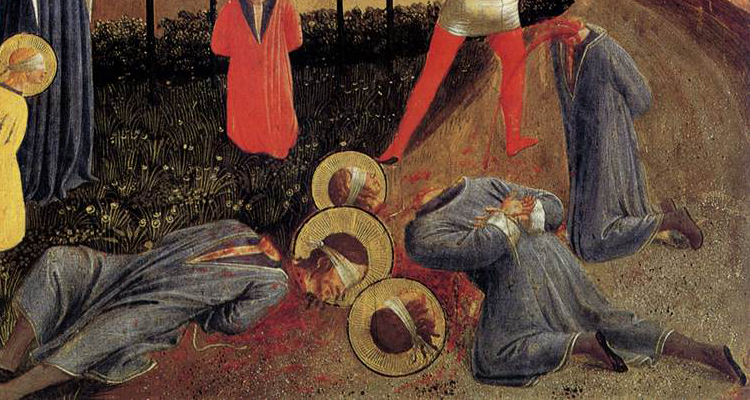}
  \caption{}
  \label{fig:image-annotation-example-3d}
\end{subfigure}%


\begin{subfigure}[t]{.05\linewidth}
  \centering
  \rotatebox{90}{\hspace{-22pt}\parbox{2.25cm}{\scriptsize\centering Positioning of \\the human body}}
\end{subfigure}%
\hfill
\begin{subfigure}{.225\linewidth}
  \centering
  \includegraphics[width=\linewidth]{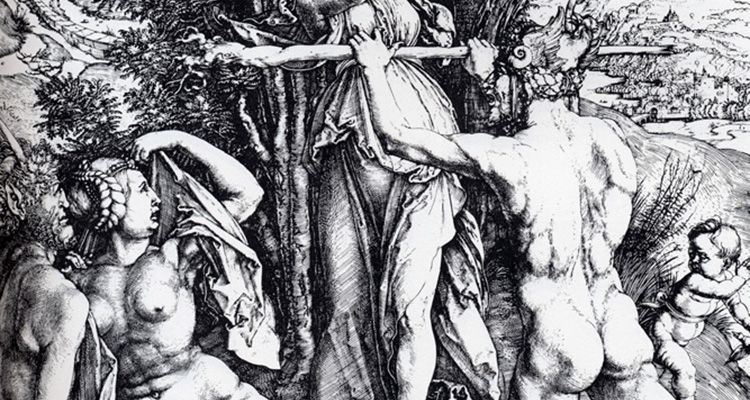}
  \caption{}
  \label{fig:image-annotation-example-4a}
\end{subfigure}%
\hfill
\begin{subfigure}{.225\linewidth}
  \centering
  \includegraphics[width=\linewidth]{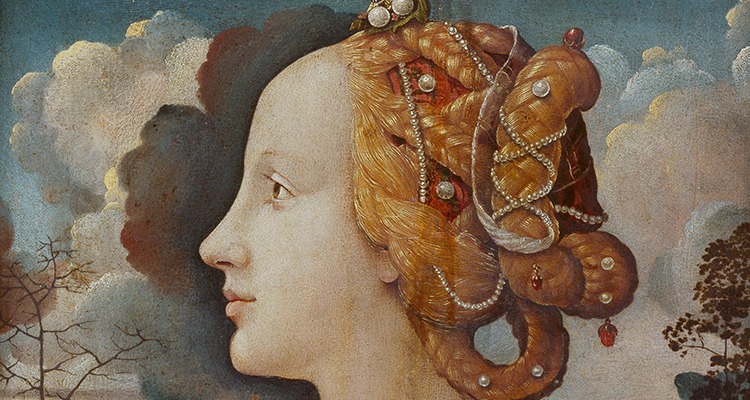}
  \caption{}
  \label{fig:image-annotation-example-4b}
\end{subfigure}%
\hfill
\begin{subfigure}{.225\linewidth}
  \centering
  \includegraphics[width=\linewidth]{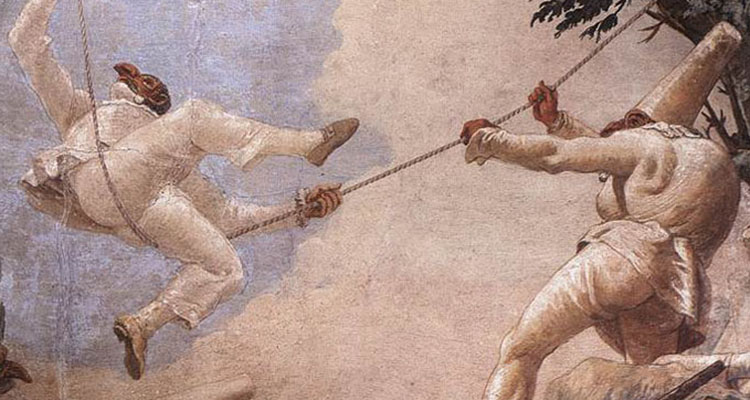}
  \caption{}
  \label{fig:image-annotation-example-4c}
\end{subfigure}%
\hfill
\begin{subfigure}{.225\linewidth}
  \centering
  \includegraphics[width=\linewidth]{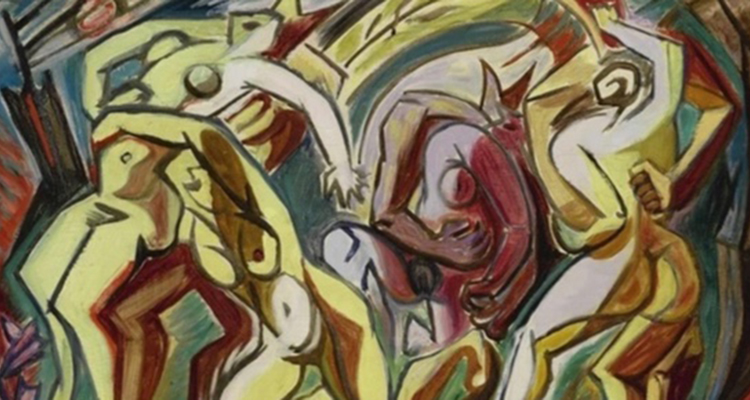}
  \caption{}
  \label{fig:image-annotation-example-4d}
\end{subfigure}%
\caption{
Sample images of the \popart\ data set illustrate the four aspects that pose challenges to the annotation of art-historical imagery: (i)~the size of human figures, (ii)~their relation to each other, (iii)~deviations from the \enquote*{ideal} human body, and (iv)~the positioning of the body. All images are in the public domain.
}
\label{fig:image-annotation-examples}
\end{figure}

\subsubsection{Exhaustiveness}
\label{chp:completeness}

We set the following guidelines to guarantee exhaustive annotation. (i)~All human-appearing figures are enclosed by bounding boxes; the distance to the outline of the human figure is to be kept as small as possible. Only the visible area of the figure is labeled and not the estimated total extent of it. Larger numbers of people, whose individual figures can no longer be sufficiently differentiated, are labeled as \enquote*{crowd.} In contrast to the Microsoft \ac{COCO}~\cite{lin2014} and PASCAL \ac{VOC} data sets~\cite{everingham2010}, we do not indicate truncated or difficult-to-annotate figures as such. (ii)~Up to four human figures per image are fine-granularly labeled with keypoints, selecting those whose limbs can be captured best. We do not consider it beneficial to label all figures with keypoints, as this would favor styles that feature an above-average number of figures---and thus would introduce data bias. Keypoints are recorded in a \enquote*{person-centric} way, i.e., left points refer to the figure's left extremities. Since in many cases keypoints are not clearly visible or are occluded, we establish three rules. (a)~If an occluded body part can be approximated by another, it is denoted by a keypoint; e.g., an elbow obscured by a pillar is annotated if the hand and shoulder of the respective body half are visible. (b)~Due to the low variance of the body parts, eyes and ears are labeled in profile views on the non-visible side of the face as well. (c)~If several joints are not visible and cannot be approximated, the corresponding keypoints are not set.

\subsubsection{Consistency}
\label{chp:consistency}

To ensure consistency in the annotation, a fixed team of annotators was employed at the Ludwig Maximilian University of Munich throughout the entire period. Annotation guidelines were discussed with the annotators prior to annotation and iteratively modified during the annotation procedure, e.g., when unusual figure constellations occurred more frequently. In the course of the process, recurring challenges arose for both modes, bounding box and keypoint annotation; \ref{fig:image-annotation-taxonomy} visualizes them in taxonomic form. We identify four major challenges: (i)~those resulting from variations of the size of human figures, (ii)~those emerging from the relation of human figures to each other, (iii)~those attributable to deviations from the \enquote*{ideal} human body, and (iv)~those originating from the body's positioning in the image space. 

\paragraph{Variations of the size of human figures.}
Large crowds and figures in the background complicate the annotation. Both cases are dominated by very small figures (\ref{fig:image-annotation-example-1a}; \ref{fig:image-annotation-example-1b}), figures that are difficult to separate from each other (\ref{fig:image-annotation-example-1c}), or that are difficult to recognize as human (\ref{fig:image-annotation-example-1d}). The latter is due not only to factors intrinsic to the object, i.e., the analog original, but also to image-extrinsic factors, i.e., the original's digital reproduction. In particular, compression artifacts or low-quality and out-of-date resolutions hamper the process.

\paragraph{Relation of human figures to each other.}
We distinguish two kinds of figure relations, which are crucial for annotation: non-referential and referential ones. Referential relations include constellations in which the body of one and the same figure is represented several times but in different ways. In addition to shadows (\ref{fig:image-annotation-example-2a}), these mainly include reflections, e.g., in mirrors (\ref{fig:image-annotation-example-2b}), in water (\ref{fig:image-annotation-example-2c}), and on surfaces like metallic armor. 
We set the corresponding bounding boxes whenever the referencing part, the reflection, can be recognized as human-like even without the referenced part, i.e., the human reflected in some way. Non-referential relations are found when figures overlap, intersect, or are symmetrically arranged (\ref{fig:image-annotation-example-2d}). In case of overlaps and intersections, we approximate occluded keypoints as far as possible.

\paragraph{Deviations from the \enquote*{ideal} human body.} 
The ideal human body has been studied since antiquity~\cite{schoen1920, speich1957, frings1998, rathgeber2019}: 
from scholars like Vitruvius~\cite{zoellner2004}, to medieval draftsmen such as Villard de Honnecourt~\cite{hahnloser1972}, Renaissance artists Leonardo da Vinci~\cite{keele1983, kemp2006, pieper2018} and Albrecht Dürer~\cite{rupprich1966, hinz2011}, or even modernists like Oskar Schlemmer~\cite{schlemmer1969}. We declare bodies as deviating from this ideal whose depicted measurements or proportions do not adhere to usual conventions. Three subcategories are discerned.
(a)~Often deviations are due to stylistic reasons expressed regionally, epochally, or individually. The lack of differentiation of the entire body shape or individual body parts is characteristic of Impressionism and Pointillism (\ref{fig:image-annotation-example-3a});
figures veiled by robes that fundamentally obscure the body are common in Art Nouveau as well as Japanese woodblock prints of the Ukiyo-e~\cite{zatlin1997}. If the placement of keypoints in an image is complicated by blurred contours, distorted proportions, or missing joints, as shown in \ref{fig:image-annotation-example-3b}, we approximate them, provided the figures can be recognized as human.
(b)~Another subcategory comprises non-human bodies and body parts. These include mythological figures such as centaurs, harpies, and mermaids (\ref{fig:image-annotation-example-3c}), biblical figures, e.g., angels, and human-like animals like monkeys and lemurs. While animals are excluded from annotation, we annotate human parts of mythological and biblical figures; consequently, the animal limbs of centaurs are not annotated, nor are the wings and halo of angels.
(c)~The third subcategory covers non-living bodies and body parts, with a considerable portion being severed heads (\ref{fig:image-annotation-example-3d})\footnote{See iconographies such as David and Goliath, Judith and Holofernes, and Salomé and John the Baptist.} and limbs. The latter may again result from the analog original itself, for instance, as part of the composition, but may also be grounded in the digital reproduction, e.g., in particularly detailed views or images in need of restoration that no longer permit keypoints to be fully labeled. While severed limbs are not annotated, severed heads are, since they generally allow for more keypoints and constitute a more substantial part of the human body than hands or legs. Also included are fabricated bodies, such as dolls, masks, crafts, sculptures, and images within images depicting human bodies, e.g., in salon paintings.

\begin{table}[t!]
\footnotesize
\begin{xltabular}{\columnwidth}{@{}XX*{7}{R}@{}}
\caption{
The \peopleart~\cite{westlake2016} and \popart\ data sets are descriptively compared. Figures are indicated by bounding boxes associated with them. Up to 17 keypoints are stored per figure. Difficult-to-annotate figures are included.
}
\label{fig:statistics-popart}\\
\toprule
Data set & Split & Images & $\text{Images}_{Pos}$ & $\text{Images}_{Neg}$ & Figures & Crowds & Keypoints & Styles\\
\midrule
\peopleart\ & Training & 1,623 & 525 & 1,098 & 1,512 & 0 & 0 & 43\\
 & Validation & 1,383 & 442 & 941 & 1,219 & 0 & 0 & 43\\
 & Testing & 1,616 & 522 & 1,094 & 1,137 & 0 & 0 & 43\\
\cmidrule{2-9}
 & Total & 4,622 & 1,489 & 3,133 & 3,868 & 0 & 0 & 43\\
\midrule
\popart\ & Training & 1,472 & 1,472 & 0 & 6,457 & 245 & 33,582 & 22\\
 & Validation & 491 & 491 & 0 & 2,175 & 114 & 11,104 & 22\\
 & Testing & 491 & 491 & 0 & 2,117 & 106 & 11,468 & 22\\
\cmidrule{2-9}
 & Total & 2,454 & 2,454 & 0 & 10,749 & 465 & 56,154 & 22\\
\bottomrule
\end{xltabular}
\end{table}

\paragraph{Positioning of the human body.}
Of relevance is the body's positioning in the image space especially for back views, as in Dürer's \textit{Hercules} (1498; \ref{fig:image-annotation-example-4a}), where the inversion of keypoints must be taken into account. 
For profile views, it is crucial to set the eyes and ears on the non-visible side of the face as well. This applies, e.g., to Florentine portraits (\ref{fig:image-annotation-example-4b}), which refer to the strict profile of emperors on ancient coins~\cite{christiansen2011}. In a considerable number of images, perspective distortions are furthermore present, along with twists and turns. They are found primarily in Baroque and Rococo works such as those by Tiepolo (\ref{fig:image-annotation-example-4c}), 
but also in \nth{20}-century avant-garde movements, as in the French surrealist André Masson (\ref{fig:image-annotation-example-4d}). 
There, too, the possible inversion of keypoints has to be considered. If a figure is twisted to such an extent that keypoints cannot be approximated, individual limbs are omitted and annotated only up to the last keypoint visible or to be approximated.

\subsubsection{Accuracy}
\label{chp:accuracy}

Figure instance annotations were checked in several test cycles according to formerly stated guidelines. They were once again reviewed at the end of the annotation process. We verified, e.g., that each keypoint referred to the correct body part, that body halves were properly labeled, especially for back views and twisted figures, and that bounding boxes surrounded only the extent of the figure visible in the image.

\begin{figure}
\centering
\begin{subfigure}{.49\linewidth}
  \centering
  \includegraphics[width=\linewidth]{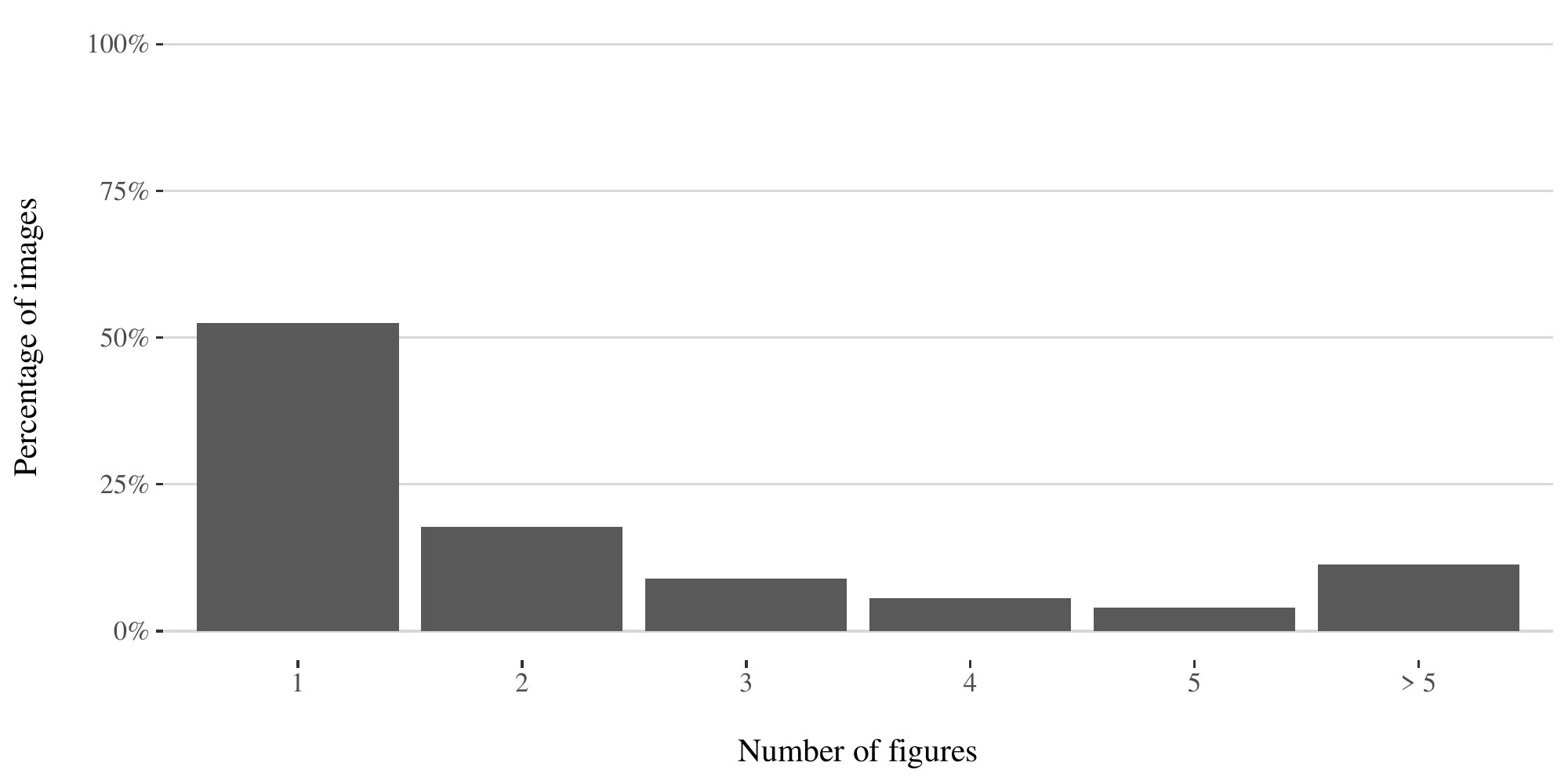}
  \caption{\peopleart}
  \label{fig:number-of-persons-people-art}
\end{subfigure}%
\hfill
\begin{subfigure}{.49\linewidth}
  \centering
  \includegraphics[width=\linewidth]{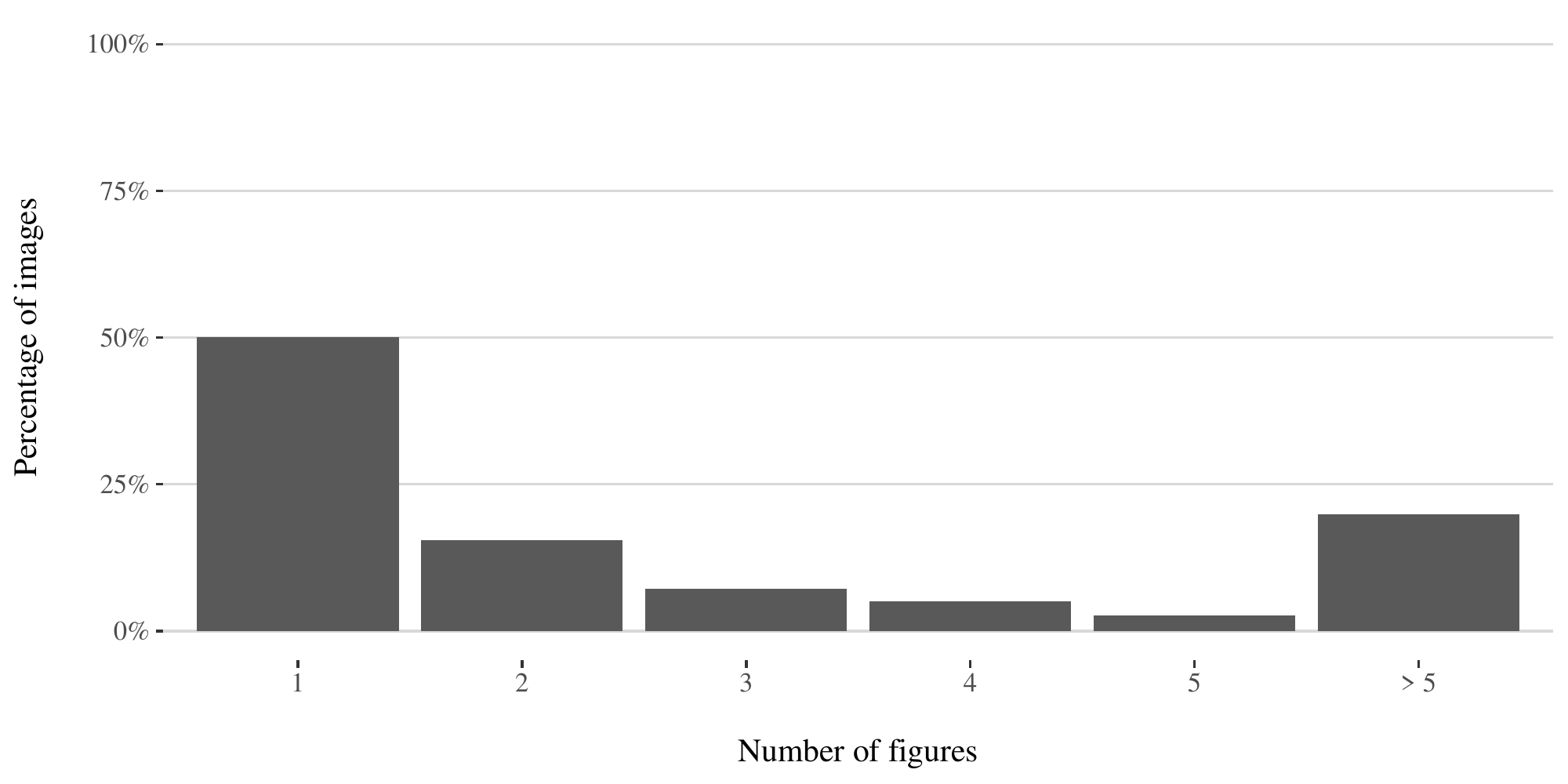}
  \caption{\popart}
  \label{fig:number-of-persons-popart}
\end{subfigure}%
\caption{
The proportion of images with at least six figures is $8.52$\,\% higher in the \popart\ than in the \peopleart\ data set~\cite{westlake2016}. This is also reflected in a larger maximum number of figures in an image: it is $28$ for \peopleart\ and $110$ for \popart.
}
\label{fig:number-of-persons}
\bigskip
\centering
\begin{subfigure}{.49\linewidth}
  \centering
  \includegraphics[width=\linewidth]{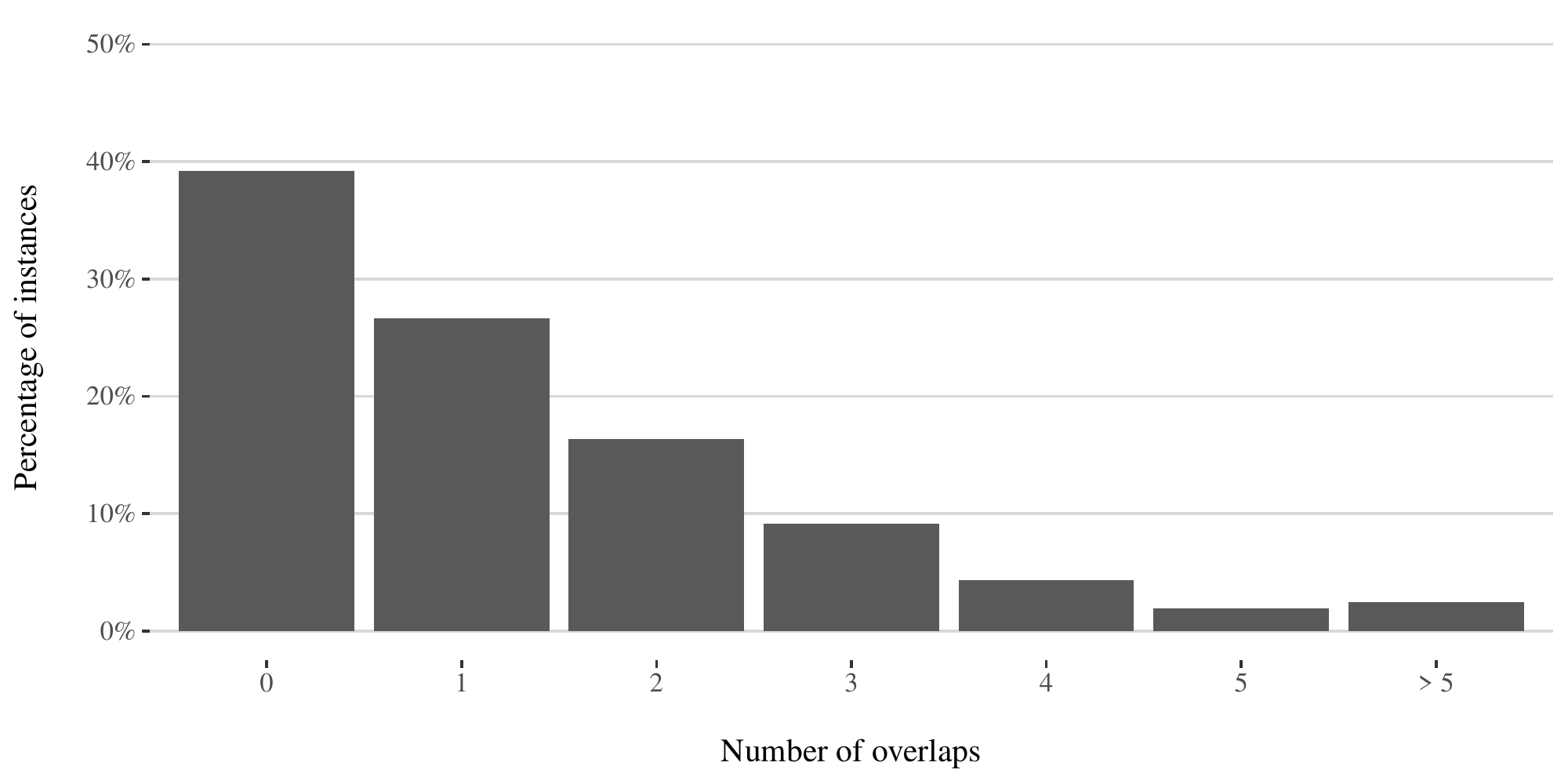}
  \caption{\peopleart}
  \label{fig:number-of-overlaps-people-art}
\end{subfigure}%
\hfill
\begin{subfigure}{.49\linewidth}
  \centering
  \includegraphics[width=\linewidth]{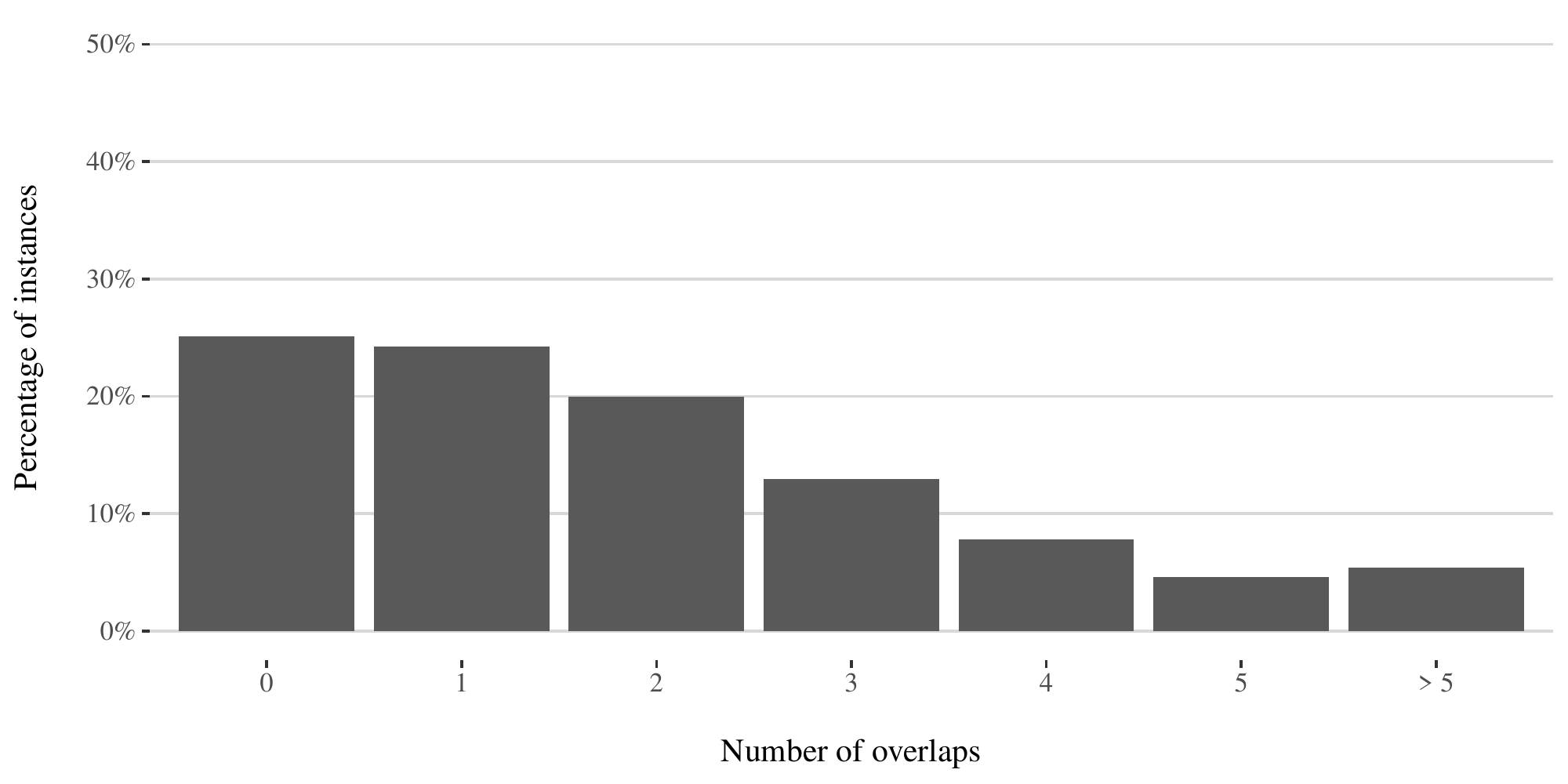}
  \caption{\popart}
  \label{fig:number-of-overlaps-popart}
\end{subfigure}%
\caption{
The \popart\ data set has a $14.22$\,\% higher share of overlapping figure instances than the \peopleart\ data set~\cite{westlake2016}, with the maximum number of overlaps in an image being $11$ for \peopleart\ and $23$ for \popart.
}
\label{fig:number-of-overlaps}
\end{figure}

\subsection{Descriptive Statistics}
\label{chp:descriptive-statistics}

For machine learning purposes, the \popart\ data set is divided into three subsets: training, validation, and testing. They contain $1{,}472$, $491$, and $491$ images, respectively, so approximate split ratios of 60\,\%, 20\,\%, and 20\,\% are met. In contrast to the \peopleart\ data set~\cite{westlake2016}, we do not reduce image sizes to a maximum scale of $500 \times 500$ pixels, but directly redistribute the digital reproductions from \wikiart. The widest image measures $6{,}298 \times 3{,}049$ and the highest $4{,}524 \times 6{,}018$ pixels. Figure instance annotations total $6{,}457$ in \popart\ for training, $2{,}175$ for validation, and $2{,}117$ for testing, with keypoint annotations of $33{,}582$, $11{,}104$, and $11{,}468$, respectively. 
To maintain an equal distribution of figure instances across data splits, we applied the following procedure: images were first grouped by depiction style and then sorted in descending order based on the number of figure instance annotations. Considering the split ratios, we then processed batches of five images and randomly assigned three of them as training samples, one as validation, and one as testing.

\begin{figure}
\centering
\begin{lstlisting}[language=json]
{
    (*@ $\cdots$ @*)
    "images": [
        (*@ $\cdots$ @*)
        {
            "id": 58,
            "license": 1,
            "width": 2310,
            "height": 3000,
            "file_name": "albrecht-durer_death-of-orpheus-1498.jpg",
            "metadata": {
                "wikiart_url": "https://www.wikiart.org/en/albrecht-durer/death-of-orpheus-1498",
                "wikiart_image_url": "https://uploads.wikiart.org/images/albrecht-durer/death-of-orpheus-1498.jpg",
                "wikiart_style": "Northern Renaissance"
            }
        },
        (*@ $\cdots$ @*)
    ],
    "annotations": [
        (*@ $\cdots$ @*)
        {
            "id": 64,
            "image_id": 58,
            "category_id": 1,
            "area": 1319355,
            "bbox": [ 616.0, 1718.0, 1305.0, 1011.0 ],
            "segmentation": [ [ 1921.0, 1718.0, 1921.0, 2729.0, 616.0, 2729.0, 616.0, 1718.0 ] ],
            "keypoints": [ 950, 1872, 2, 945, 1848, 2, 904, 1870, 2, 924, 1873, 2, 864, 1918, 2, 1108,
                1905, 2, 871, 2100, 2, 1279, 1778, 2, 790, 2339, 2, 1085, 1789, 2, 750, 2620, 2, 1313,
                2311, 2, 1147, 2339, 2, 1600, 2567, 2, 902, 2629, 2, 1870, 2456, 2, 1200, 2431, 2 ],
            "num_keypoints": 17,
            "iscrowd": false
        },
        (*@ $\cdots$ @*)
    ]
}
\end{lstlisting}
\caption{
\popart\ follows the JSON-based Microsoft COCO format~\cite{lin2014}, for which a figure instance annotation with its referencing image annotation is displayed.
}
\label{fig:coco-format}
\end{figure}

\ref{fig:statistics-popart} summarizes \popart\ in comparison to the similarly constituted \peopleart\ data set. 
Both data sets focus exclusively on human figures depicted in art-historical objects; other classes are not annotated. While \peopleart's core application solely lies in the computer-aided detection of figures, \popart\ is designed to support both their detection and that of their keypoints. Further structural differences arise.
(i)~\peopleart\ contains more training, validation, and testing images due to the integration of negative image samples that do not show human figures. However, \popart\ features almost three times as many positive samples in the training set, which have at least one instance annotation. This includes more small-area instances measuring between $0$ and $16^2$ pixels, namely $0.9$\,\%. In the \peopleart\ data set, it amounts to only $0.5$\,\% despite reduced image sizes. 
In addition, the data set completely lacks crowd annotations. \popart\ thus decisively enables the automatic detection even of figures that are displayed small.
(ii)~Moreover, \popart\ accounts for the broad spectrum of art-historical body language in two ways. As shown in \ref{fig:number-of-persons}, the proportion of images with at least six figures is $8.52$\,\% higher in the \popart\ than in the \peopleart\ data set. This is reflected in a larger maximum number of figures in an image: it is $28$ for \peopleart\ and $110$ for \popart. As a result, the \popart\ data set also has a $14.22$\,\% higher share of overlapping figure instances than \peopleart\ (\ref{fig:number-of-overlaps}), with the maximum number of overlaps in an image being $11$ for \peopleart\ and $23$ for \popart.

\subsection{Data Split Format}
\label{chp:data-split-format}

All data splits follow the JSON-based Microsoft \ac{COCO} format~\cite{lin2014}; \ref{fig:coco-format} displays an exemplary figure instance annotation with its referencing image annotation. For each image annotation, we provide metadata (\texttt{wikiart\_url}, \texttt{wikiart\_image\_url}, and \texttt{wikiart\_style}) in addition to mandatory fields (\texttt{id}, \texttt{license}, \texttt{width}, \texttt{height}, and \texttt{file\_name}). Figure instance annotations contain task-agnostic information (\texttt{id}, \texttt{image\_id}, and \texttt{category\_id}), supplemented by fields essential to the respective detection task. The fields \texttt{bbox}, \texttt{segmentation}, and \texttt{iscrowd} are declared for both figure instance and keypoint detection, while \texttt{keypoints} and \texttt{num\_keypoints} are noted for keypoint detection only. Through a 53-dimensional array, each of the $17$ keypoints is represented with three values: its location, $x$ and $y$, and a visibility flag $v$ that indicates whether the respective keypoint is visible and labeled, $v=2$, or not, $v=0$. In contrast to the Microsoft \ac{COCO} format guidelines, we assign $v=2$ to index occluded keypoints as well, rather than $v=1$. Keypoints are recorded in the order established by the \ac{COCO} format: nose, left and right eye, left and right ear, left and right shoulder, left and right elbow, left and right wrist, left and right hip, left and right knee, left and right ankle.

\section{Applications}
\label{chp:applications}

In the course of this section, we consider application scenarios in which \popart\ can be usefully integrated and, building on these, discuss prospects for a digitally supported art history. Two scenarios are distinguished: those arising from human pose estimation (Section~\ref{chp:human-pose-estimation}), and those from human figure detection (Section~\ref{chp:human-detection}).

\begin{figure}
\centering
\includegraphics[width=\linewidth]{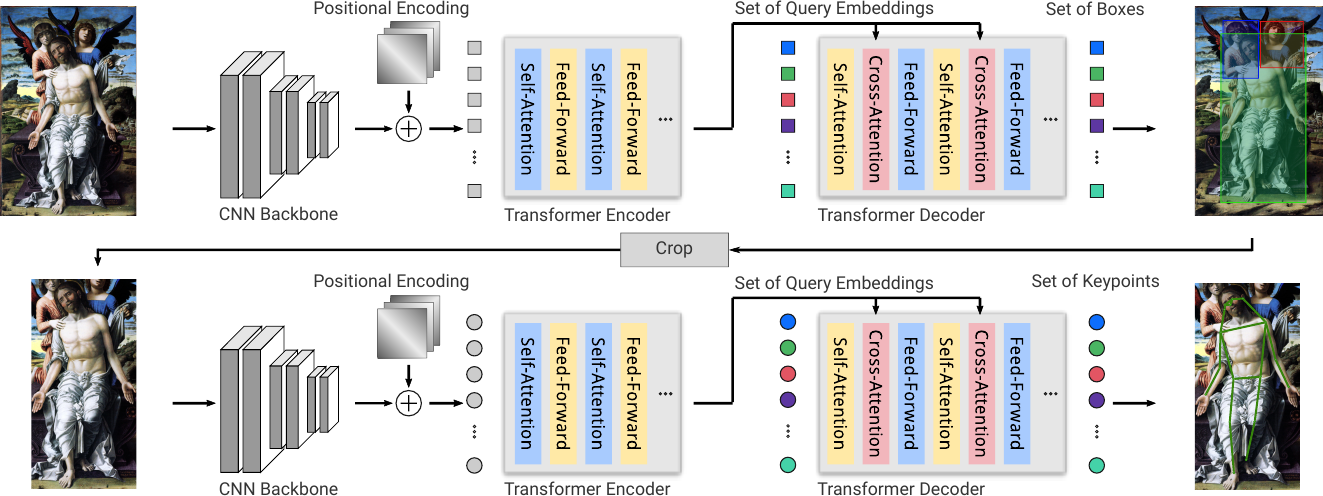}
\caption{
The two-stage human pose estimator from \citet{springstein2022} uses two Transformer models: the input of the first stage is the entire image, for which the first Transformer predicts a fixed set of bounding boxes. The individual boxes are cropped and serve as input for the second stage; the second Transformer model then computes a set of keypoints.
}
\label{fig:two-stage-human-pose-estimator}
\end{figure}
\begin{figure}
\centering
\begin{subfigure}{.49\linewidth}
  \centering
  \includegraphics[width=\linewidth]{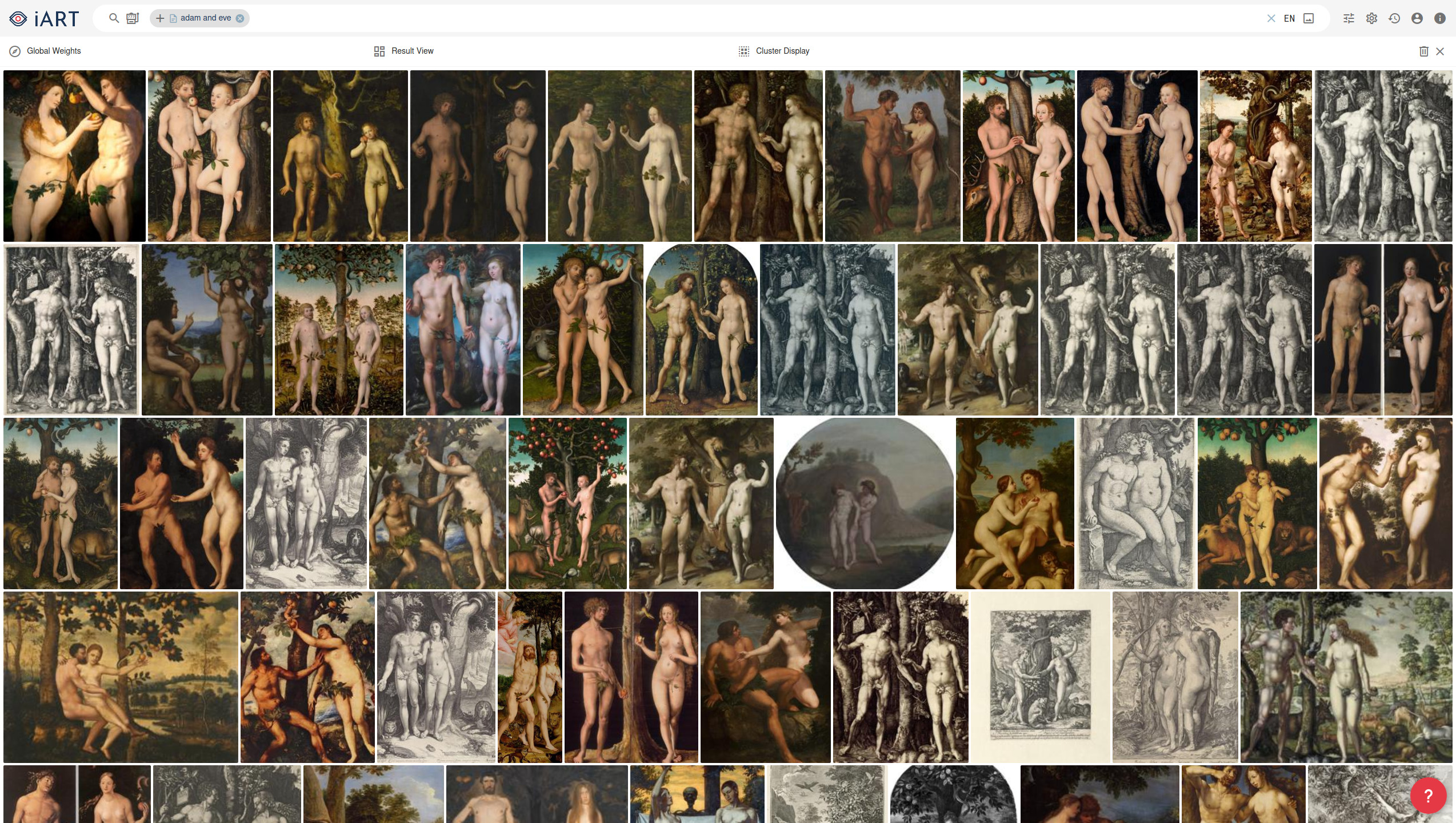}
  \caption{Default image grid}
  \label{fig:iart-image-grid}
\end{subfigure}%
\hfill
\begin{subfigure}{.49\linewidth}
  \centering
  \includegraphics[width=\linewidth]{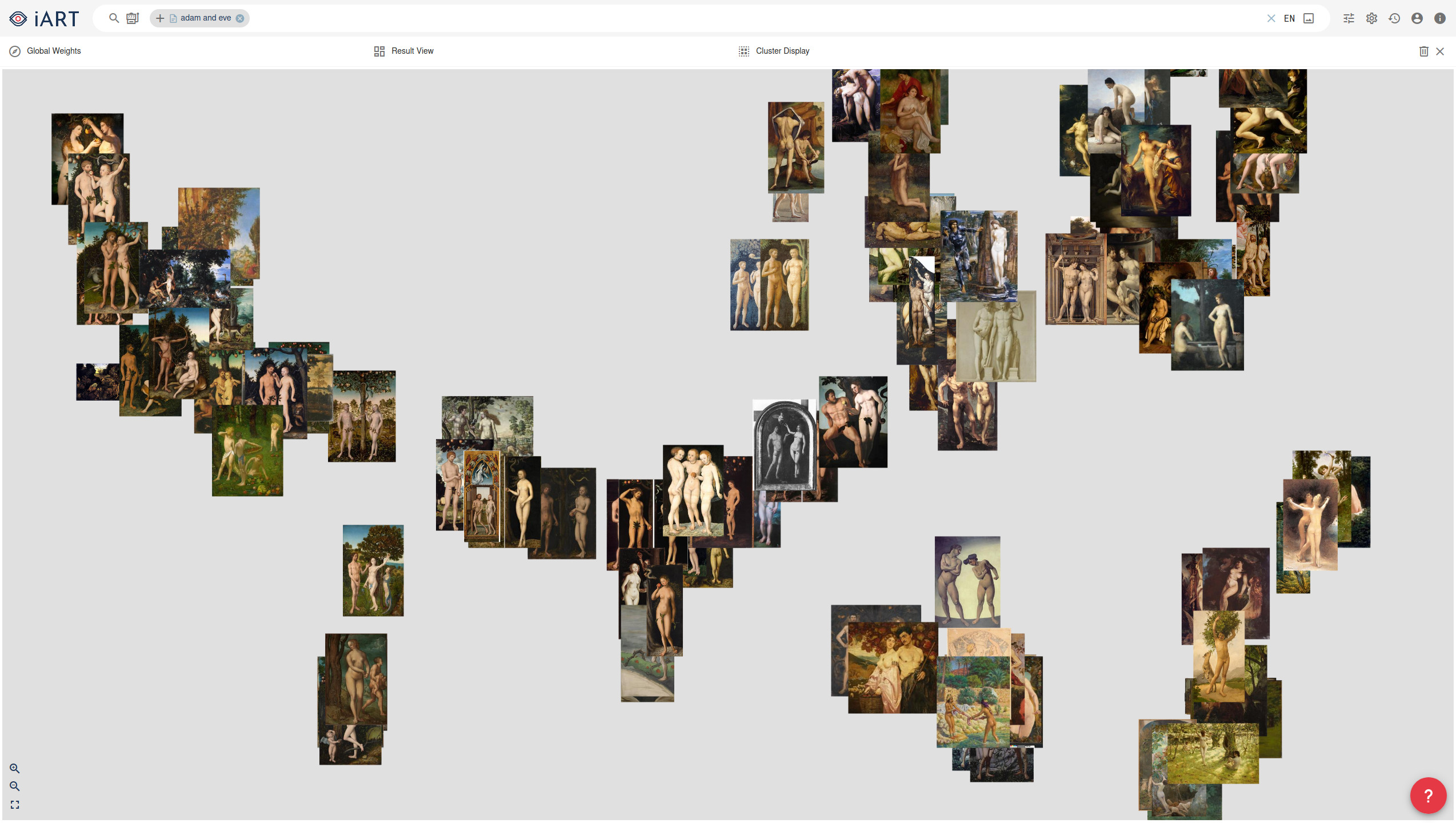}
  \caption{Two-dimensional canvas view}
  \label{fig:iart-canvas-view}
\end{subfigure}%
\caption{
With the aid of the web platform \iart~\cite{springstein2021, schneider2022}, the process of comparative vision is facilitated by various object views, as illustrated by the example of \textit{Fall of Man}.
}
\label{fig:iart}
\end{figure}

\subsection{Human Pose Estimation}
\label{chp:human-pose-estimation}

In a first application scenario, we demonstrate that \popart\ enables the quantitatively systematized exploration of human poses in visual art. For this purpose, we suggested in \citet{springstein2022} a two-stage approach based on two Transformer models~\cite{vaswani2017, carion2020}: the first model detects bounding boxes of human figures, while the second one analyzes the individual boxes for keypoints (\ref{fig:two-stage-human-pose-estimator}). We in this context adapted a semi-supervised learning technique to reduce the performance loss caused by the shift between existing real-world data sets and the art-historical domain, and to reduce the quantity of domain-specific annotation data. The basic principle is to use both labeled and unlabeled image material to train a student model. The teacher serves as a generator of pseudo-labels; to this end, unlabeled images are first weakly augmented and then used for the detection of human figures, just as figures enclosed by bounding boxes are weakly augmented and used to predict their keypoints. Three art-historical data sets are plugged into the routine: in addition to \popart, we also employ \peopleart~\cite{westlake2016} for labeled and Art500k~\cite{art500k} for unlabeled data. Experiments performed on the \popart\ test set in comparison to more established approaches that apply pre-trained models~\cite{jenicek2019, madhu2020b} or enrich real-world data sets with style transfer~\cite{madhu2020a} indicated that the performance of human pose estimators is greatly enhanced by using semi-supervised methods with additional unlabeled data. Moreover, in a user study, we also confirmed the feasibility of the approach for retrieval tasks, enabling the search for resembling poses. The pose---as the holistic abstraction of bodily expression---can thus prove elemental to the formulaic recapitulation of significant motifs through computational assistance.

This becomes particularly evident when machine-generated similarity arrangements are explored through web-based user interfaces. For instance, on the platform \iart~\cite{springstein2021, schneider2022},\footnote{\url{https://www.iart.vision/}.} object retrieval is performed not only based on art-historical keywords generated by deep learning, but also by leveraging state-of-the-art multimodal embeddings such as the Transformer-backed neural network \acsu{CLIP}, which creates a unified feature space for image and text~\cite{radford2021}. First, the retrieval of certain iconographies is thereby enabled. As illustrated in \ref{fig:iart-image-grid}, searching for \enquote{adam and eve} primarily returns the classical Renaissance depiction of the \textit{Fall of Man}, in which Adam and Eve stand image-parallel, left and right under the Tree of Knowledge. The iconography can be examined more in-depth if, on top of \ac{CLIP}-based pre-filtering, the pose embeddings of each figure are determined and then mapped onto a two-dimensional canvas using the dimensionality reduction technique \acs{UMAP}~\cite{umap}. Several cluster structures emerge in \ref{fig:iart-canvas-view}: the one shown at the top left, e.g., reveals an image group of more dynamic poses that are conspicuous for their bent or flared legs; apart from the fact that here the apple is being handed to Adam in a rather prominent manner.


\begin{table}[t!]
\footnotesize
\begin{xltabular}{\columnwidth}{@{}XlT*{7}{S}@{}}
\caption{
Figure detection results are reported for the \peopleart\ test set~\cite{westlake2016}. For training and validation, \popart\ was used in addition to \peopleart. In contrast to previous benchmarks by \citet{kadish2021} and \citet{gonthier2022}, we include difficult-to-annotate figures. The best performing approach is indicated in bold.
}
\label{tab:detection-benchmark-people-art-popart}\\
\toprule
Model & Backbone & LR & $\text{AP}$ & $\text{AP}_{50}$ & $\text{AP}_{75}$ & $\text{AP}_{S}$ & $\text{AP}_{M}$ & $\text{AP}_{L}$ & $\text{AR}$\\
\midrule
\ac{TOOD} \cite{tood} & ResNet-50-FPN & $2\textrm{e}-4$ & 0.478 & 0.780 & 0.499 & \textbf{0.162} & 0.311 & 0.511 & \textbf{0.654}\\
\ac{PVT} \cite{pvt} & PVTv2-B2 & $1\textrm{e}-5$ & \textbf{0.497} & \textbf{0.805} & \textbf{0.518} & 0.076 & \textbf{0.315} & \textbf{0.532} & 0.625\\
Cascade R-CNN \cite{cascade_rcnn} & ResNet-50-FPN & $2\textrm{e}-4$ & 0.464 & 0.761 & 0.490 & 0.152 & 0.307 & 0.495 & 0.606\\
SABL Cascade R-CNN \cite{sabl_faster_rcnn} & ResNet-50-FPN & $2\textrm{e}-4$ & 0.456 & 0.762 & 0.457 & 0.116 & 0.311 & 0.487 & 0.601\\
Faster R-CNN \cite{faster_rcnn} & ResNet-50-FPN & $2\textrm{e}-4$ & 0.439 & 0.770 & 0.447 & 0.128 & 0.312 & 0.465 & 0.580\\
SABL Faster R-CNN \cite{sabl_faster_rcnn} & ResNet-50-FPN & $2\textrm{e}-4$ & 0.453 & 0.756 & 0.463 & 0.129 & 0.308 & 0.483 & 0.604\\
PISA Faster R-CNN \cite{pisa_faster_rcnn} & ResNet-50-FPN & $2\textrm{e}-4$ & 0.447 & 0.767 & 0.464 & 0.133 & 0.306 & 0.475 & 0.582\\
Libra Faster R-CNN \cite{libra_faster_rcnn} & ResNet-50-FPN & $2\textrm{e}-4$ & 0.442 & 0.769 & 0.451 & 0.084 & 0.312 & 0.471 & 0.583\\
\bottomrule
\end{xltabular}
\end{table}

\subsection{Human Figure Detection}
\label{chp:human-detection}

We show in the second application scenario that as a by-product of \popart's domain-specific curation, the sole detection of art-historical figures is decisively improved. For this purpose, we utilize the same models and pipeline as described in Section~\ref{chp:image-collection} for the preliminary step of our semi-automatic image collection procedure: models are first pre-trained on Microsoft \ac{COCO} 2017 for $12$ epochs and then fine-tuned, with their classification head re-initialized, for another $12$ epochs---now on both the \peopleart~\cite{westlake2016} and \popart\ data sets. Parameter settings remain unchanged; the learning rate is, again, set to $2\textrm{e}-4$ in case of ResNet-50 and $1\textrm{e}-5$ in case of Transformer backbones. Compared to those in \ref{tab:detection-benchmark-people-art}, the benchmarks shown in \ref{tab:detection-benchmark-people-art-popart} clearly demonstrate that \ac{AP} and \ac{AR} increase considerably for all models when \popart\ is integrated into the training routine. For the Transformer-based \ac{PVT} model \cite{pvt}, e.g., \ac{AP} and \ac{AR} improve to the same extent, from $46.5$ to $49.7$\,\% and $60.1$ to $62.5$\,\%, respectively. The leap is even more noticeable if we plug-in the \popart\ instead of the \peopleart\ test set. \ac{AP} then rises from $36.6$ to $43.6$\,\% and \ac{AR} from $48.2$ to $55.0$\,\% for \ac{PVT}. At the same time, this reconfirms the greater complexity of the figures contained in \popart, which are exhaustively marked in the images by bounding boxes, even if they are very small or appear in crowds, and hence overlap frequently. The additional integration of \popart\ into the training routine thus is particularly advantageous to movements that emphasize the depiction of a larger number of people, as in Mannerism and the regional expressions of the Renaissance; in Northern Renaissance works, e.g., \ac{AP} improves from $27.1$ to $33.5$\,\% and \ac{AR} from $37.5$ to $43.4$\,\% (\ref{tab:detection-benchmark-people-art-popart-style} in Appendix).

\begin{figure}
\centering
\begin{subfigure}{.49\linewidth}
  \centering
  \includegraphics[width=\linewidth]{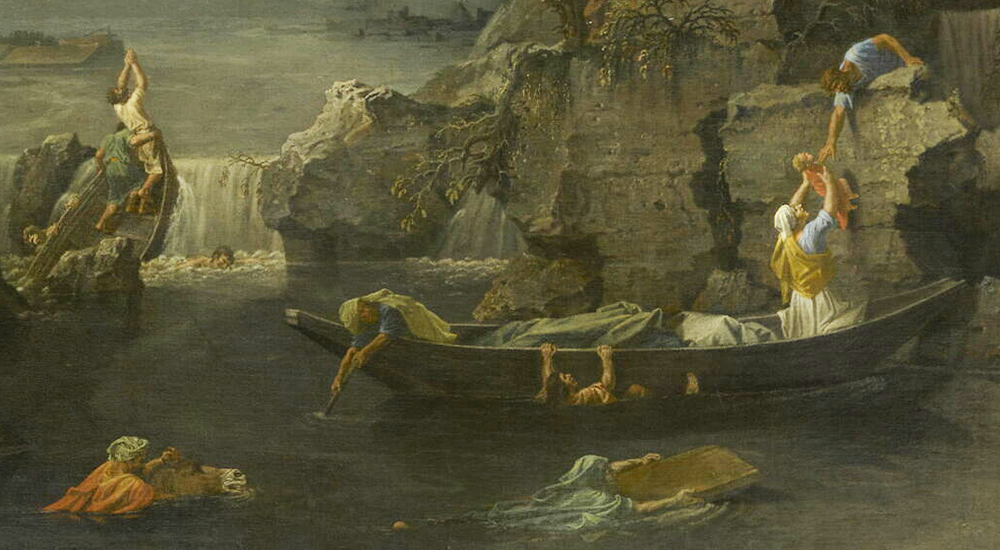}
  \caption{Nicholas Poussin, \textit{The Deluge} (1660--1664)}
  \label{fig:crowd-poussin}
\end{subfigure}%
\hfill
\begin{subfigure}{.49\linewidth}
  \centering
  \includegraphics[width=\linewidth]{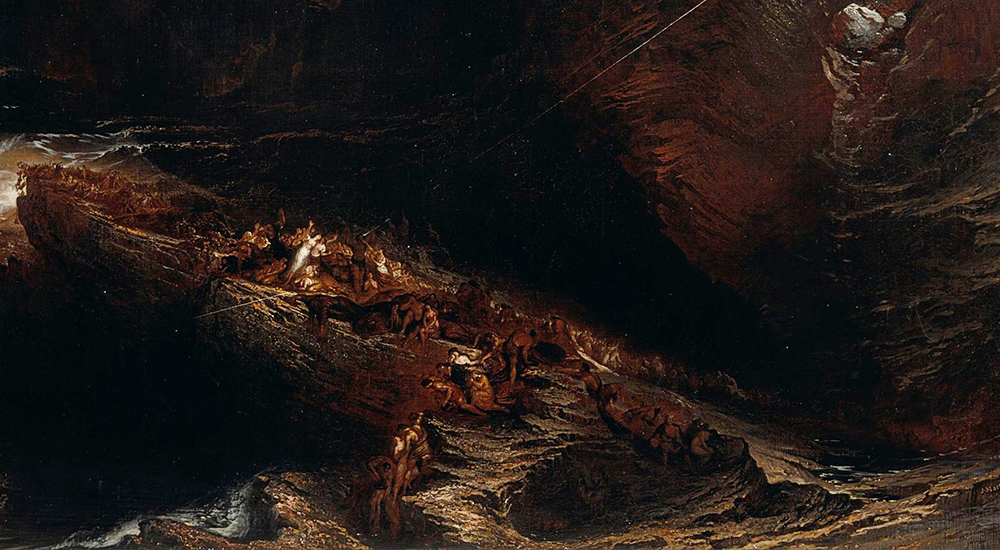}
  \caption{John Martin, \textit{The Deluge} (1834)}
  \label{fig:crowd-martin}
\end{subfigure}%
\caption{
Detail views of the crowds depicted in Nicholas Poussin's and John Martin's versions of \textit{The Deluge}, respectively. Both images have been slightly lightened to emphasize depiction specifics. The images are in the public domain.
}
\label{fig:crowd-depictions}
\end{figure}

Indeed, this image of the crowd, from small gatherings in village squares to streams of passers-by in modern pedestrian zones, benefits especially from computer-aided methods of detection; even if these may initially only be used to pre-filter the (digitally available) image material. Namely, the crowd's underlying constitution, which has been increasingly received since the \nth{18} century \cite{kemp1973}, becomes strictly quantifiable: by the number of people in it, their proximity or distance from each other, the space they occupy in the image, and in relation to other subjects. John Martin's emphatically apocalyptic \textit{Deluge} (1834; \ref{fig:crowd-martin}), for instance, focuses on the entirely de-individualized crowd---a multitude of people depicted in a confined space, who are \enquote{tossed back and forth like cue balls} \cite{kohle2007}. In Nicholas Poussin's \textit{Deluge} (1660--1664; \ref{fig:crowd-poussin}), on the other hand, the majority of figures are still, because of the larger body size, differentiated in their moments of action. While a man clings to his horse in the foreground, a mother, slightly moved back, stretches her child upwards to the shore. It is precisely these iconographic traditions that first become easily decipherable in larger amounts of data through distant viewing and only then are examined in detail from a more art-historical perspective. Recommender systems like \iart~\cite{springstein2021, schneider2022} can ultimately point to research-worthy phenomena here as well.

\section{Conclusion}
\label{chp:conclusion}

In this paper, we introduced with \textit{Poses of People in Art} the first publicly available and openly licensed data set for estimating human poses in visual art. It consists of $2{,}454$ images from $22$ art-historical depiction styles, including those that increasingly turned away from lifelike representations of the body and toward artificial forms. A total of $10{,}749$ human figures are enclosed by rectangular bounding boxes, with a maximum of four per image labeled by up to $17$ keypoints. For machine learning purposes, the data set is pre-split into three subsets---training, validation, and testing---, each following the JSON-based Microsoft \ac{COCO} format. In addition to mandatory fields, image annotations provide metadata from the art-historical online encyclopedia \wikiart. As illustrated in two application scenarios, the data set not only validates the performance of deep-learning models, but in this way enables the comprehensive investigation of body phenomena in art---whether at the level of individual figures, whose bodily subtleties are captured, or entire figure constellations, whose position, distance, or proximity to one another is considered. With the further aid of readily accessible online platforms like the presented \iart, we see the potential to reveal large-scale disruptions of formal conventions and make them interactively explorable. Since this would allow hitherto marginalized collections to be easily included in analyses, the discipline of art history would benefit from an increasingly de-canonized gaze that is no longer primarily devoted to European art. Intra- as well as inter-iconographic recurrent motifs, whose radically altered semantics are disconcerting, might be thoroughly discussed for the first time in this context.

\begin{acks}
This work was funded in part by the German Research Foundation (Deutsche Forschungsgemeinschaft, DFG) under project no. 415796915. We thank Ursula Huber for her valuable support with the image annotation. We also thank Hubertus Kohle, Ralph Ewerth, and Matthias Springstein for fruitful discussions and useful comments on the subject matters.
\end{acks}

\section*{Authors' Contributions}
S.S. conceived, designed, and performed the experiments, analyzed the data, oversaw image annotation, and ensured data quality; R.V. performed image annotation. S.S. and R.V. wrote the manuscript, and read, commented, and approved the final version.

\bibliographystyle{ACM-Reference-Format}
\bibliography{library}

\appendix

\section*{Appendix}
\label{appendix}

\footnotesize

\begin{xltabular}{\columnwidth}{@{}Xl*{7}{S}@{}}
\caption{
Figure detection results are reported for the \popart\ test set by depiction style and training set(s); \ac{TOOD}~\cite{tood} is employed as model, respectively. In contrast to previous benchmarks by \citet{kadish2021} and \citet{gonthier2022}, we include difficult-to-annotate figures. 
}
\label{tab:detection-benchmark-people-art-popart-style}\\
\toprule
Style & Training Set(s) & $\text{AP}$ & $\text{AP}_{50}$ & $\text{AP}_{75}$ & $\text{AP}_{S}$ & $\text{AP}_{M}$ & $\text{AP}_{L}$ & $\text{AR}$\\
\midrule
\endhead
Abstract Expressionism & \peopleart\ & 0.900 & 1.000 & 1.000 & & & 0.900 & 0.900\\
 & \peopleart, \popart\ & 0.850 & 1.000 & 1.000 & & & 0.850 & 0.850\\
\midrule
Art Nouveau & \peopleart\ & 0.425 & 0.677 & 0.441 & & 0.228 & 0.449 & 0.643\\
 & \peopleart, \popart\ & 0.460 & 0.762 & 0.452 & & 0.222 & 0.486 & 0.722\\
\midrule
Baroque & \peopleart\ & 0.301 & 0.461 & 0.322 & 0.000 & 0.024 & 0.444 & 0.386\\
 & \peopleart, \popart\ & 0.357 & 0.539 & 0.389 & 0.051 & 0.047 & 0.512 & 0.498\\
\midrule
Contemporary Realism & \peopleart\ & 0.624 & 0.864 & 0.724 & 0.316 & 0.501 & 0.730 & 0.720 \\
 & \peopleart, \popart\ & 0.627 & 0.851 & 0.746 & 0.255 & 0.632 & 0.755 & 0.729\\
\midrule
Cubism & \peopleart\ & 0.511 & 0.836 & 0.565 & & 0.750 & 0.511 & 0.696\\
 & \peopleart, \popart\ & 0.601 & 0.876 & 0.655 & & 0.676 & 0.607 & 0.752\\
\midrule
Early Renaissance & \peopleart\ & 0.427 & 0.696 & 0.442 & 0.000 & 0.178 & 0.541 & 0.563\\
 & \peopleart, \popart\ & 0.503 & 0.774 & 0.548 & 0.000 & 0.295 & 0.605 & 0.629\\
\midrule
Expressionism & \peopleart\ & 0.567 & 0.832 & 0.594 & & 0.486 & 0.627 & 0.711\\
 & \peopleart, \popart\ & 0.592 & 0.839 & 0.627 & & 0.474 & 0.667 & 0.754\\
\midrule
Fauvism & \peopleart\ & 0.493 & 0.765 & 0.534 & & 0.126 & 0.579 & 0.622\\
 & \peopleart, \popart\ & 0.576 & 0.845 & 0.643 & & 0.171 & 0.657 & 0.690\\
\midrule
High Renaissance & \peopleart\ & 0.254 & 0.405 & 0.268 & 0.000 & 0.052 & 0.424 & 0.340\\
 & \peopleart, \popart\ & 0.310 & 0.481 & 0.319 & 0.002 & 0.068 & 0.514 & 0.417\\
\midrule
Impressionism & \peopleart\ & 0.473 & 0.737 & 0.489 & 0.000 & 0.341 & 0.520 & 0.616\\
 & \peopleart, \popart\ & 0.509 & 0.777 & 0.527 & 0.000 & 0.397 & 0.549 & 0.656\\
\midrule
Mannerism & \peopleart\ & 0.298 & 0.540 & 0.283 & 0.000 & 0.069 & 0.397 & 0.483\\
 & \peopleart, \popart\ & 0.371 & 0.668 & 0.343 & 0.022 & 0.192 & 0.459 & 0.542\\
\midrule
Naive Art & \peopleart\ & 0.291 & 0.470 & 0.304 & 0.166 & 0.150 & 0.396 & 0.443\\
 & \peopleart, \popart\ & 0.394 & 0.679 & 0.379 & 0.175 & 0.279 & 0.487 & 0.528\\
\midrule
New Realism & \peopleart\ & 0.514 & 0.803 & 0.538 & & 0.557 & 0.521 & 0.665\\
 & \peopleart, \popart\ & 0.553 & 0.842 & 0.547 & & 0.373 & 0.593 & 0.716\\
\midrule
Northern Renaissance & \peopleart\ & 0.271 & 0.460 & 0.286 & 0.015 & 0.128 & 0.410 & 0.375\\
 & \peopleart, \popart\ & 0.335 & 0.555 & 0.350 & 0.052 & 0.196 & 0.481 & 0.434\\
\midrule
Pointillism & \peopleart\ & 0.465 & 0.726 & 0.556 & 0.000 & 0.467 & 0.519 & 0.567\\
 & \peopleart, \popart\ & 0.553 & 0.827 & 0.644 & 0.010 & 0.570 & 0.600 & 0.638\\
\midrule
Pop Art & \peopleart\ & 0.454 & 0.628 & 0.499 & 0.041 & 0.258 & 0.555 & 0.559\\
 & \peopleart, \popart\ & 0.514 & 0.683 & 0.580 & 0.164 & 0.351 & 0.600 & 0.667\\
\midrule
Post Impressionism & \peopleart\ & 0.607 & 0.903 & 0.660 & & 0.396 & 0.628 & 0.732\\
 & \peopleart, \popart\ & 0.672 & 0.911 & 0.704 & & 0.445 & 0.693 & 0.768\\
\midrule
Realism & \peopleart\ & 0.657 & 0.869 & 0.768 & 0.000 & 0.047 & 0.727 & 0.746\\
 & \peopleart, \popart\ & 0.693 & 0.915 & 0.721 & 0.030 & 0.347 & 0.756 & 0.787\\
\midrule
Rococo & \peopleart\ & 0.534 & 0.787 & 0.595 & & 0.023 & 0.601 & 0.629\\
 & \peopleart, \popart\ & 0.606 & 0.866 & 0.654 & & 0.184 & 0.654 & 0.710\\
\midrule
Romanticism & \peopleart\ & 0.303 & 0.499 & 0.306 & 0.000 & 0.098 & 0.436 & 0.414\\
 & \peopleart, \popart\ & 0.401 & 0.619 & 0.440 & 0.000 & 0.176 & 0.557 & 0.508\\
\midrule
Symbolism & \peopleart\ & 0.322 & 0.574 & 0.316 & 0.068 & 0.228 & 0.360 & 0.458\\
 & \peopleart, \popart\ & 0.362 & 0.674 & 0.362 & 0.069 & 0.276 & 0.403 & 0.521\\
\midrule
Ukiyo-e & \peopleart\ & 0.414 & 0.746 & 0.437 & 0.000 & 0.050 & 0.460 & 0.619\\
 & \peopleart, \popart\ & 0.437 & 0.830 & 0.429 & 0.009 & 0.080 & 0.479 & 0.638\\
\bottomrule
\end{xltabular}

\end{document}